\documentclass[10pt,journal,compsoc]{IEEEtran}
\usepackage{amsmath,amssymb,amsfonts}
\usepackage{algorithmic}
\usepackage{graphicx}
\usepackage{epstopdf}
\usepackage{textcomp}
\usepackage{xcolor}
\usepackage{diagbox}
\usepackage{threeparttable}
\usepackage{booktabs}
\usepackage{siunitx}
\usepackage{float}
\usepackage{bm}
\usepackage{diagbox}
\usepackage{array}
\usepackage{booktabs}
\usepackage{makecell}
\usepackage{caption}
\usepackage{subcaption}
\usepackage{framed,multirow}
\usepackage{fancyhdr}

\newcommand{\tabincell}[2]{\begin{tabular}{@{}#1@{}}#2\end{tabular}}

\usepackage{ifpdf}

\ifCLASSOPTIONcompsoc
  \usepackage[nocompress]{cite}
\else
  \usepackage{cite}
\fi
\ifCLASSINFOpdf
\else
\fi
\begin{document}
%
\title{A Softmax-free Loss Function Based on Predefined Optimal-distribution of\\ Latent Features for Deep Learning Classifier}
%
%
%
%

\author{Qiuyu Zhu,
        Xuewen Zu
\thanks{\emph{(Corresponding author: Qiuyu Zhu)}}
\IEEEcompsocitemizethanks{\IEEEcompsocthanksitem Q. Y. Zhu is with the School of Communication and Information Engineering, Shanghai University, Shanghai 200444, China.\protect\\
E-mail: zhuqiuyu@staff.shu.edu.cn
\IEEEcompsocthanksitem X. W. Zu is with the School of Communication and Information Engineering, Shanghai University, Shanghai 200444, China.\protect\\
E-mail: xuewenzu@shu.edu.cn
}
}
%
%

\markboth{Journal of \LaTeX\ Class Files,~Vol.~14, No.~8, August~2015}%
{Shell \MakeLowercase{\textit{et al.}}: Bare Demo of IEEEtran.cls for Computer Society Journals}
%



\IEEEtitleabstractindextext{%
\begin{abstract}
In the field of pattern classification, the training of deep learning classifiers is mostly end-to-end learning, and the loss function is the constraint on the final output (posterior probability) of the network, so the existence of Softmax is essential. In the case of end-to-end learning, there is usually no effective loss function that completely relies on the features of the middle layer to restrict learning, resulting in the distribution of sample latent features is not optimal, so there is still room for improvement in classification accuracy. Based on the concept of Predefined Evenly-Distributed Class Centroids (PEDCC), this article proposes a Softmax-free loss function based on predefined optimal-distribution of latent features—POD Loss. The loss function only restricts the latent features of the samples, including the norm-adaptive Cosine distance between the latent feature vector of the sample and the center of the predefined evenly-distributed class, and the correlation between the latent features of the samples. Finally, Cosine distance is used for classification. Compared with the commonly used Softmax Loss, some typical Softmax related loss functions and PEDCC-Loss, experiments on several commonly used datasets on several typical deep learning classification networks show that the classification performance of POD Loss is always significant better and easier to converge. Code is available in https://github.com/TianYuZu/POD-Loss.
\end{abstract}

}

\maketitle




\IEEEdisplaynontitleabstractindextext

%
\IEEEpeerreviewmaketitle

\IEEEraisesectionheading{\section{Introduction}\label{sec:introduction}}

%
%
%
%
\IEEEPARstart{I}{n} recent years, convolutional neural networks (CNN) have achieved great success in image classification \cite{10.1145/3065386}, semantic segmentation \cite{7298965}, face recognition \cite{electronics9081188}, target tracking \cite{7780834} and target detection \cite{ZAFEIRIOU20151}. The algorithm based on convolutional neural networks has become one of the leading technologies in these fields. In order to further improve the performance of convolutional neural networks, researchers start with the network structure, from AlexNet \cite{10.1145/3065386} to VGGNet \cite{Simonyan2015VeryDC}, and then to the deeper ResNet \cite{7780459}, DenseNet \cite{8099726}, etc. Many effective solutions have also been put forward in other aspects such as data enhancement, batch normalization \cite{10.5555/3045118.3045167} and various activation functions.

Loss function is an indispensable part of neural network model. It assists in updating the parameters of neural network model during the training phase. The traditional Softmax loss function is composed of softmax plus cross-entropy loss function. Because of its advantages of fast learning speed and good performance, it is widely used in image classification. However, Softmax loss function adopts an inter-class competition mechanism, only cares about the accuracy of the prediction probability of the correct label, ignores the difference of the incorrect label, and can not ensure the compactness of intra-class and the discreteness of inter-class.  L-Softamax \cite{10.5555/3045390.3045445} adds an angle constraint on the basis of Softmax to ensure that the boundaries of samples of different classes are more obvious. A-Softmax \cite{8100196} also improves Softamx, and proposes weights normalized and angular spacing to realize the recognition standard that the maximum intra-class distance is smaller than the minimum intra-class distance. In addition, AM-Softmax \cite{8331118} further improves A-Softmax. In order to improve the convergence speed, Euclidean feature space is converted to Cosine feature space, and $cos(m\theta)$ is changed to $(cos\theta - m)$. Center Loss \cite{10.1007/978-3-319-46478-7_31} adds the constraints of sample features before the classification layer for the first time on the basis of Softmax (the mean square error between the sample features and the calculated class centers is used to restrict the intra-class distance and the inter-class distance), but the distance between similar classes cannot be well separated.

\begin{figure*}
  \centering
  \includegraphics[width=0.9\textwidth]{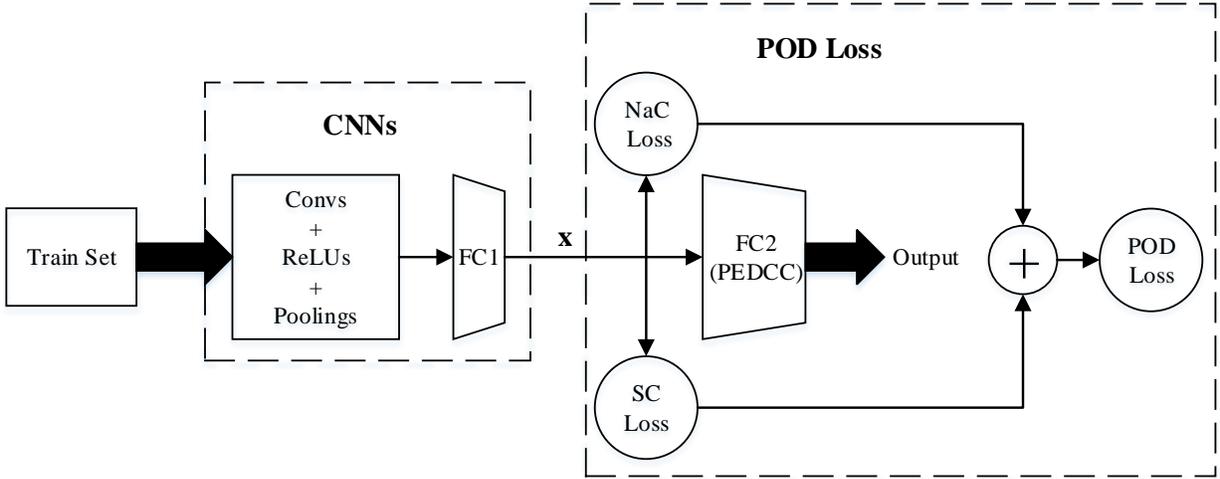}
\caption{POD Loss of CNN classifier. POD Loss is a combination of Norm-adaptive Cosine (NaC) Loss and Self-correlation Constraint (SC) Loss, \textbf{x} is the latent features of the sample, FC2 (PEDCC) is the linear classification layer with parameter solidification. POD Loss is completely a constraint on the optimal distribution of latent features of the samples.}
\end{figure*}

There are also many other loss functions \cite{2019arXiv190301182C}\cite{7258343}\cite{9511448} to study the other shortcomings of softmax, and achieved certain results. Nevertheless, because convolutional neural networks mostly use end-to-end learning, most of the research direction of loss function is the constraint on the final output of the network, which is a posteriori probability, and there are few constraints after the feature extraction layer. From the perspective of features, if we want to get the features with the greatest discreteness of inter-class and the best compactness of intra-class, it is a better method to restrict the features directly and uniquely. PEDCC-Loss \cite{2019arXiv190200220Z}\cite{8933403} predefines evenly-distributed class centroids to replace the continuously updated class centers in Center Loss, so as to maximize the inter-class distance. Meanwhile, using AM-Softmax \cite{8331118} and mean square error (MSE) loss to restrict the compactness of feature distribution of intra-class and the discreteness of feature distribution of inter-class. PEDCC-Loss improves the classification accuracy in both face recognition and general image classification, but still contains the constraint of the loss function of the posterior probability, which makes the sample feature distribution still not in the optimal state.

This article proposes a Softmax-free loss function (POD Loss) based on predefined optimal-distribution of latent features for deep learning classifier. The proposed POD Loss discards the constraints on the posterior probability in the traditional loss function, and only restricts the extracted sample features to achieve the optimal distribution of latent feature, i.e. maximization of inter-class distance and minimization of intra-class distance. On the one hand, the predefined evenly-distributed class centroids (PEDCC) is used, and the weight of the classification linear layer in the network (the weight is fixed in the training phase to maximize the inter-distance) is replaced, so as to maximize inter-class distance in latent features. On the other hand, Norm-adaptive Cosine (NaC) Loss restricts the Cosine distance between the sample feature vector and the PEDCC, resulting in the minimization of intra-class distance. At the same time, the correlations between the sample features are limited, so that the extracted latent features are the most effective. Finally, the Cosine distance is used for classification to obtain high classification accuracy. 

The location of POD Loss and the whole network structure are shown in Fig. 1. Section 3 gives the details of the method.

Our main contributions are as follows:
\begin{itemize}
\item The PEDCC we proposed before is adopted in our classification model, and only the output of the feature extraction layer in the neural network is restricted to achieve the optimal distribution of latent features. The Softmax loss function is discarded, the weight of the classification layer is fixed, and the Cosine distance is used for classification.
\item We find that the classification accuracy of samples with high $l_2$-norm of latent features in common CNN networks is significantly higher than that of samples with low value, and then can use this characteristic to design loss function to improve the recognition rate.
\item Based on the characteristics of PEDCC and latent feature, a norm-adaptive Cosine loss is used to constrain the Cosine distance between the sample features and the PEDCC class centroid, where an additional value is added dynamically to the $l_2$-norm of the latent feature. At the same time, the correlation between the latent feature dimensions is minimized to improve the classification accuracy.
\item For the classification tasks, experiments on multiple datasets and different networks are conducted. Compared with the traditional cross-entropy loss plus softmax, some typical Softmax related loss functions and PEDCC-Loss, the classification accuracy of POD Loss is obviously better, and the training of network is easier to converge.
\end{itemize}

This paper is arranged as follows: Related works are introduced in Section 2. In Section 3, our approach is described in detail. Afterwards, the validity of our method is verified through the experimental results on different datasets and various network structures. Finally, in Section 5, we summarize the paper and introduce future work.

\section{Related works}
\subsection{PEDCC}
Predefined Evenly-Distributed Class Centriods (PEDCC) \cite{2019arXiv190200220Z} are a group of randomly evenly-distributed points on the hyper-sphere surface, which can be obtained by physical model method \cite{2019arXiv190200220Z} and mathematical generation method \cite{9444709}. For a deep learning classifier, the distance between different classes is not optimal, that is, similar classes are close and different classes are far away, which greatly affects the classification performance. PEDCC can optimize the distance between different classes by mapping the generated evenly-distributed points to the class center of each class.

For datasets of different classes, \cite{9444709} gives the theorem: For arbitrarily generated $k$ point $a_i(i=1,2,...,k)$ evenly-distributed on the unit hypersphere of $n$ dimensional Euclidean space, if $k \leq n+1$, such that $\langle a_i, a_j \rangle=-\frac{1}{k-1}, i \neq j$. It means that $k$ evenly-distributed PEDCC class centroids can be obtained when $k$(number of classes) $\leq n$(feature dimension)$+1$ is satisfied. \cite{9444709} proposed that the $n$-dimensional space of PEDCC can be divided into two orthogonal subspaces: ($k-1$)-dimensional subspace spanned by $k$ PEDCC points and ($n-k+1$)-dimensional subsapce. Although most of the sample features finally converge to the ($k-1$)-dimensional subspace, that is, the Cosine value of the sample features and the ($k-1$)-dimensional subspace is about $1$, the ($n-k+1$)-dimensional subspace is also useful in the training process, which benefits other layers better extract features in the back propagation process.

\subsection{Loss function of classification}
In the field of classification, there are many different loss functions in neural networks for end-to-end learning. For multi-classification problems, Softmax loss function is generally selected, which has good performance and is easy to converge. The Softmax loss function is as follows:
\begin{normalsize}
\begin{equation}
L_{Softmax}=\frac{1}{N}\sum_{i} -log{\frac{e^{z_{y_i}}}{\sum_{j}e^{z_j}}}
\end{equation}
\end{normalsize}
where $N$ represents the number of samples, $z_{y_i}$ represents the output value of the last fully connected layer of the correct class $y_i$, and $z_j$ is the output value of the last fully connected layer of the $j$-th class.

Softmax loss mainly uses the information from the correct label to maximize the possibility of data, but largely ignores the information from the remaining incorrect labels. Complement Objective Training (COT) \cite{2019arXiv190301182C} believes that in addition to correct labels, incorrect labels should also be used in training, which can effectively improve the performance of the model. The training strategy is to alternate training between the correct label and the incorrect labels. Even considering the information of adding incorrect labels, the promotion of COT has a certain limit. Orthogonality Loss \cite{9184823} considers enlarging the inter-class variance by directly penalizing weight vectors of last fully connected layer, which represent the center of classes. The main idea is that in order for weight vectors to be discriminative, it should be as close as possible to be orthogonal to each other in the vector space. However, its application scope is only for face recognition, and there are doubts about the effect of other classification tasks. 

From the perspective of maximize inter-class distance and minimize intra-class distance, Balanced Separate Loss (BS Loss) \cite{9511448} based on the normalized Cosine distance can learn more discriminative features. Centralized Large Margin Cosine Loss (C-LMCL) \cite{8666165} combines Large Margin Cosine Loss (LMCL) \cite{10.5555/3045390.3045445}\cite{8331118} and Center Loss \cite{10.1007/978-3-319-46478-7_31}. The C-LMCL considers that, for LMCL, the feature vectors of different classes are far enough away in the feature space, but the intra-class is not compact enough; for center loss, it focuses on pulling features of the same class as close as possible, but the distances among different classes might be not far enough. Nevertheless, the disadvantage of the above two loss functions is that they both have limitations in the scope of application. PEDCC-Loss \cite{2019arXiv190200220Z}\cite{8933403} proposes predefined evenly-distributed class centroids instead of the continuously updated class centers in center loss \cite{10.1007/978-3-319-46478-7_31}, and uses the fixed PEDCC weight instead of the weight of the classification linear layer in the neural network to maximize the inter-class distance. At the same time, the constraint of latent features is added (a constraint similar to Center Loss is added to calculate the mean square error (MSE) loss of sample features and PEDCC center), and AM-Softmax is also applied. This method makes the distribution of intra-class features more compact and the distribution of inter-class features more distant. But PEDCC-Loss still contains the constraint of a posterior probability loss function, which makes the sample feature distribution unable to be in the optimal state. The PEDCC-Loss expression is as follows:
\begin{small}
\begin{equation}
L_{PEDCC-AM}=-\frac{1}{N}\sum_{i} log{\frac{e^{{s\cdot}({\cos\theta_y}_i-m)}}{e^{{s\cdot}({\cos\theta_y}_i-m)}+\begin{matrix} \sum_{j=1,j\ne y_i}^c e^{{s\cdot}{\cos\theta_j}}\end{matrix}}}
\end{equation}
\end{small}
\begin{normalsize}
\begin{equation}
L_{PEDCC-MSE}=\frac{1}{2N}\sum_{i=1}^N{\left \| \bm{x_i}-\bm{pedcc_{y_i}} \right \|}^2
\end{equation}
\end{normalsize}
\begin{normalsize}
\begin{equation}
L_{PEDCC-Loss}=L_{PEDCC-AM}+\lambda\sqrt[n]{L_{PEDCC-MSE}}
\end{equation}
\end{normalsize}
where $N$ is the number of samples, $\lambda$ is a weighting coefficient, and $n \geq 1$ is a constraint factor of $L_{PEDCC-MSE}$. $\bm{x_i}$ represents the input feature of the ith sample (normalized), and $\bm{pedcc_{y_i}}$ represents the PEDCC central feature of the class to which the sample belongs (normalized).

The intuition behind Multi-loss Regularized Deep Neural Network \cite{7258343} is that the extra loss functions with different theoretical motivations may drag the algorithm away from overfitting to one particular single-loss function (e.g., Softmax Loss). Therefore, the multi-loss framework introduces the cross-loss-function regularization for boosting the generalization capability of deep neural network. However, the characteristic that each loss function corresponds to a fully connected layer greatly increases the amount of network parameters and calculation, and the training speed becomes slower.

\subsection{Loss function for latent features in self-supervised learning}
Self-supervised learning \cite{8014803} is one of unsupervised learning, which mainly hopes to learn a general feature expression for downstream tasks. The main method is to supervise oneself through oneself and predict the missing information by relying on the information around the missing part. 

In self-supervised learning, Barlow Twins \cite{2021arXiv210303230Z} proposes an innovative loss function, adding a decorrelation mechanism to the loss function to maximize the variability of representation learning. Barlow Twins Loss is written as:
\begin{normalsize}
\begin{equation}
L_{BT}=\sum_{i}(1-C_{ii})^2+\lambda\sum_{i}\sum_{j \neq i}C_{ij}^2
\end{equation}
\end{normalsize}
Where $\lambda$ is a weighting coefficient, which is used to weigh the two items in the loss function, and $C$ is the cross-correlation matrix of the output features of the sample and its enhanced sample under the same batch of two same network. The addition of the latter redundancy reduction term in the loss function reduces the redundancy between the network output features, so that the output features contains the non-redundant information of the sample, and achieves a better feature representation effect. 

On this basis, VICReg \cite{2021arXiv210504906B} combines the variance term with a decorrelation mechanism based on redundancy reduction and covariance regular term. The covariance criterion removes the correlation between the different dimensions of the learned representation, and the purpose is to spread information between dimensions to avoid dimension collapse. The criterion mainly punishes the off-diagonal terms of the covariance matrix, which further improves the classification performance of features.

In this article, PEDCC is also used to generate the predefined evenly-distributed class centroids, instead of the continuously updated class centers in Center Loss, and the solidified PEDCC weights replace the weights of the classification linear layer to maximize the inter-class distance. On the one hand, NaC Loss calculates and constrains the distance between the sample feature and the central features of the PEDCC. On the other hand, similar to self-supervised learning, a decorrelation mechanism is introduced. The self-correlation matrix of the difference between the sample features and the central features of PEDCC in each batch is calculated, and the correlation between any pair of features is restricted to improve the classification accuracy. 

In the task of image classification, the method in this article no longer restricts the posterior probability of neural network output, but restricts the latent features extracted from sample information, and realizes the optimal distribution of latent features for classification.

\section{Method}
\subsection{Norm-adaptive Cosine Loss}
\subsubsection{Cosine Loss}
The mean square error (MSE) loss function can be used to constrain the distance between the sample feature and its class PEDCC feature center. In this article, the PEDCC center and the sample feature vector are normalized before calculation, so the MSE Loss expression and its derivation are as follows:
\begin{normalsize}
\begin{equation}
L_{MSE}=\frac{1}{2N}\sum_{i=1}^N{\left \| \bm{\tilde{x_i}}-\bm{\tilde{pedcc_{y_i}}} \right \|}^2
\end{equation}
\end{normalsize}
\begin{normalsize}
\begin{equation}
=\frac{1}{2N}\sum_{i=1}^N( ||\bm{\tilde{x_i}}||^2+||\bm{\tilde{pedcc_{y_i}}}||^2-2\bm{\tilde{x_i}} \cdot \bm{\tilde{pedcc_{y_i}}})^2
\end{equation}
\end{normalsize}
\begin{normalsize}
\begin{equation}
=\frac{1}{N}\sum_{i=1}^N(1-cos{\theta_{y_i}})
\end{equation}
\end{normalsize}
where $N$ represents the number of samples, $\bm{\tilde{x_i}}$ is the feature vector of the i-th sample after normalized, $\bm{\tilde{pedcc_{y_i}}}$ is the corresponding PEDCC center after normalized, and $cos{\theta_{y_i}}$ is the Cosine value between $\bm{\tilde{x_i}}$ and $\bm{\tilde{pedcc_{y_i}}}$. It can be seen that the loss function is essentially a constraint on the Cosine distance between the sample feature and the center of the PEDCC. Taking $\theta_{y_i}$ as the independent variable to derive the derivative of $(1-cos{\theta_{y_i}})$, the derivative is $sin{\theta_{y_i}}$, as shown in Fig. 2. It can be seen from Fig. 2 that in the range of $0^{\circ}$ to $90^{\circ}$, the derivative value becomes larger and larger, and in the range of $90^{\circ}$ to $180^{\circ}$, the derivative value becomes smaller and smaller, and the derivative value is always less than or equal to $1$ in the whole range of $0^{\circ}$ to $180^{\circ}$.

\begin{figure}
  \centering
  \includegraphics[width=0.3\textwidth]{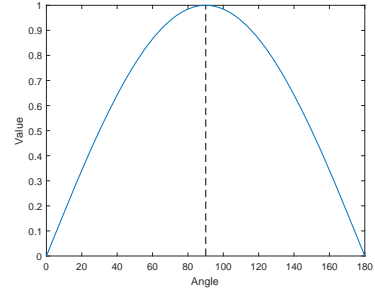}
\caption{The change of $sin{\theta_{y_i}}$ within the range of $0^{\circ}$ to $180^{\circ}.$}
\end{figure}

$\theta_{y_i}$ represents the angle between the sample feature and the predefined center feature of the $i$-th class. The angle between the sample feature point and the predefined center of the class to the origin must also be in the range of $0^{\circ}$ to $180^{\circ}$. In the process of network learning, what we want is let $\theta_{y_i}$ approach $0^{\circ}$ as quickly as possible. Therefore, the faster the falling speed of $\theta_{y_i}$, the better, that is, the greater the derivative value, the better, and $(1-cos{\theta_{y_i}})$ does not completely conform to the mathematical formula we want. Therefore, the expression of Cosine Loss is as follows:
\begin{normalsize}
\begin{equation}
L_{Cosine}=\frac{1}{N}\sum_{i=1}^N(1-cos{\theta_{y_i}})^2
\end{equation}
\end{normalsize}
where $N$ represents the number of samples, and $cos{\theta_{y_i}}$ is the Cosine value between the sample feature and the corresponding PEDCC. $\theta_{y_i}$ is used as an independent variable to derive the derivative of $(1-cos{\theta_{y_i}})^2$, and the derivative is $2\cdot(1-cos{\theta_{y_i}}) \cdot sin{\theta_{y_i}}$, as shown in Fig. 3. It can be seen from the figure that the derivative value becomes larger and larger in the range of $0^{\circ}$ to $120^{\circ}$, and the derivative value reaches $1$ around $65^{\circ}$ and the maximum value reaches about $2.5$. 
\begin{figure}
  \centering
  \includegraphics[width=0.45\textwidth]{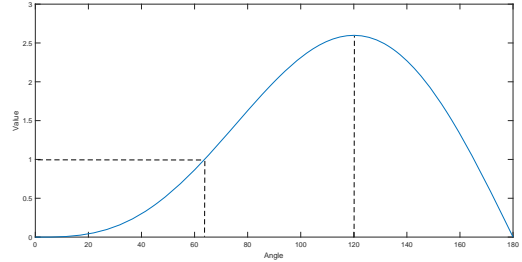}
\caption{The change of $2\cdot(1-cos{\theta_{y_i}})\cdot sin{\theta_{y_i}}$ within the range of $0^{\circ}$ to $180^{\circ}.$}
\end{figure}

\subsubsection{The influence of $l_2$-norm of latent feature}
In the experiment based on Cosine Loss (the specific implementation details are given in Section 4), we find that the classification accuracy of the samples with high $l_2$-norm of latent features in CNN networks is significantly greater than that of samples with low value. As shown in TABLE 1, low value represents the classification accuracy of samples whose $l_2$-norm value is lower than the average value of all samples, and high value represents the classification accuracy of samples whose $l_2$-norm value is higher than the average value of all samples. This phenomenon can be understood as follows: if the norm is small, it usually means that the value of features is small, and smaller features often have large relative errors, resulting in recognition errors. Then, if we can try to make the samples with lower norm close to the mean, it will not only improve the recognition accuracy, but also make the intra-class distribution of the samples more compact. 

The same rule applies to Softmax Loss. \cite{2017arXiv170510284Y} adds a feature incay on the basis of Softmax Loss. With the feature incay, feature vectors are further pushed away from the origin along the direction of their corresponding weight vectors, which achieves better inter class separability. In addition, the feature incay encourages intra-class compactness along the directions of weight vectors by increasing the small feature norm faster than the large ones. 

\begin{table}[t]
\caption{Comparison of classification accuracy (\%) of low value and high value of $l_2$-norm of latent features from the network on different dataset}
\centering
\setlength{\tabcolsep}{7mm}{
\begin{tabular}{lcc}
\toprule
  Dataset & Low Value & High Value 
  \\
\midrule
  CIFAR10 & 88.77 &  96.32 
  \\
  CIFAR100 & 71.83  & 82.23
  \\
  Tiny ImageNet & 54.65 & 69.99
  \\
  FaceScrub & 87.49 & 94.61
  \\
\bottomrule
\end{tabular}}
\end{table}

\subsubsection{Norm-adaptive Cosine Loss}
In formula 9, $cos{\theta_{y_i}}$ is generally calculated as follows:
\begin{normalsize}
\begin{equation}
cos{\theta_{y_i}}=\frac{\bm{x_i} \cdot \bm{\tilde{pedcc_{y_i}}}}{||\bm{x_i}|| \cdot ||\bm{\tilde{pedcc_{y_i}}}||+\varepsilon}
\end{equation}
\end{normalsize}
\begin{normalsize}
\begin{equation}
=\frac{\bm{x_i} \cdot \bm{\tilde{pedcc_{y_i}}}}{||\bm{x_i}||+\varepsilon}
\end{equation}
\end{normalsize}
where $\bm{x_i}$ is the sample feature before normalized, $\bm{\tilde{pedcc_{y_i}}}$ is the corresponding PEDCC center after normalized, and $\varepsilon$ is added to the denominator to avoid the denominator being equal to $0$, which is generally taken as the fixed value $10^{-12}$.

The results in subsection 3.1.2 show that the classification accuracy of the samples with high $l_2$-norm of latent features is obviously greater than that of the samples with low value. We believe that there is still room for improvement of the samples with low value, and Cosine Loss needs to be modified to improve the lower accuracy of the samples with low value while maintaining the higher accuracy of the samples with high value. The preliminary assumption is to replace $\varepsilon$ in formula 11 with a larger value $\delta$:
\begin{normalsize}
\begin{equation}
cos'{\theta_{y_i}}=\frac{\bm{x_i} \cdot \bm{\tilde{pedcc_{y_i}}}}{||\bm{x_i}||+\delta}
\end{equation}
\end{normalsize}
\begin{normalsize}
\begin{equation}
=\frac{||\bm{x_i}|| \cdot ||\bm{\tilde{pedcc_{y_i}}}|| \cdot cos{\theta_{y_i}} }{||\bm{x_i}||+\delta}
\end{equation}
\end{normalsize}
\begin{normalsize}
\begin{equation}
=\frac{||\bm{x_i}|| \cdot cos{\theta_{y_i}} }{||\bm{x_i}||+\delta}
\end{equation}
\end{normalsize}
\begin{normalsize}
\begin{equation}
=\frac{cos{\theta_{y_i}}}{1 + \frac{\delta}{||\bm{x_i}||}}
\end{equation}
\end{normalsize}
From the analysis of formula 9 and formula 15, in order to achieve the minimum possible loss, that is, the value of formula 15 approaches $1$, the improved loss function promotes the latent feature $\bm{x_i}$ in CNN networks to learn in two directions. One direction is that the latent feature $\bm{x_i}$ approximates the corresponding PEDCC class center as much as possible to make the numerator $cos{\theta_{y_i}}$ in formula 15 as close as possible to $1$, which is also the learning direction before the improvement, but our method will make the approximation constraint more strict and further reduce the intra-class variance. The other direction is to increase the $||\bm{x_i}||$ in the denominator, which is the $l_2$-norm of latent features, because the increase of $||\bm{x_i}||$ can offset the impact of the introduction of $\delta$, making the value of formula 15 approach to $1$. For the same value of $\delta$, the smaller the $||\bm{x_i}||$ is, the greater the impact of $\delta$ will be. So the improvement can selectively increase the $l_2$-norm of latent features of the samples with low value.

Many factors have been considered on how to choose the value of $\delta$. At the beginning of training, the value of $\delta$ needs to be set to a low value to prevent interfering with the process of latent features of the sample approaching the center of the class. At the later training, the value of $\delta$ needs to be set to a high value to increase the $l_2$-norm value of latent features. Therefore, the value of $\delta$ is associated with the epoch of training. In order to selectively increase the $l_2$-norm of latent features of the samples with low value, without affecting the accuracy of the samples with high value, and make the $l_2$-norm of latent features increase adaptively in the training process, the additional value $\delta$ is also related with the mean value of $l_2$-norm of all training samples. Finally, the improved $cos'{\theta_{y_i}}$, which is named as $cos_{Na}{\theta_{y_i}}$, is as follows:
\begin{normalsize}
\begin{equation}
cos_{Na}{\theta_{y_i}}=\frac{cos{\theta_{y_i}}}{1 + \frac{\delta(e,M)}{||\bm{x_i}||}}
\end{equation}
\end{normalsize}
$\delta(e,M)$ is the adaptive additonal value, that is $\delta(e,M) = \alpha \cdot e \cdot M$, where $\alpha$ is the weighted coefficient, $e$ represents the $e$-th epoch of training, $M$ is the average $l_2$-norm of latent features of all training samples after the previous epoch of training and its initial value is set to 0.05.

Based on above, Norm-adaptive Cosine (NaC) Loss is proposed, whose expression is as follows:
\begin{normalsize}
\begin{equation}
L_{NaC}=\frac{1}{N}\sum_{i=1}^N(1-cos_{Na}{\theta_{y_i}})^2
\end{equation}
\end{normalsize}
where $N$ represents the number of samples, and $cos_{Na}{\theta_{y_i}}$ represents the improved $cos{\theta_{y_i}}$.

\subsection{Selection of latent feature dimension}
For a given type of dataset, \cite{9444709} gives the theorem: For arbitrarily generated $k$ point $a_i(i=1,2,...,k)$ evenly-distributed on the unit hypersphere of $n$ dimensional Euclidean space, if $k \leq n+1$, such that $\langle a_i, a_j \rangle=-\frac{1}{k-1}, i \neq j$. Corresponding to the neural network, $k$ evenly-distributed PEDCC class centroids can be obtained when $k$(number of classes) $\leq n$(feature dimension)$+1$ is satisfied.

Different feature representations can be regarded as different knowledge of the images. The more comprehensive the knowledge, the better the classification performance, and the multi-feature dimension is indeed easier to find a hyperplane to separate images of different classes. However, too many feature dimensions will cause the classifier to emphasize the accuracy of the training set too much, and even learn some wrong or abnormal data, resulting in over-fitting problems \cite{824819}. Therefore, under the premise of $k \leq n+1$, selecting the appropriate feature dimension is also crucial to the classification performance. Through a large number of experiments, the optimal range of feature dimension $n$ is between $2*k$ and $40*k$, Section 4 gives the final selection details between the number of classes and the dimension of features.

\subsection{Decorrelation between latent features}
The features of predefined evenly-distributed class centroids are irrelevant, but the sample features during the training process only approximate the features of the class centroids to which they belong, which can not be completely equal. Therefore, there is always a certain correlation between the features. Meanwhile, under the premise of ensuring that the number of center points $k$ (the number of classes) and the space dimension $n$ satisfy $k \leq n+1$, $n$ is generally taken greater than $k$. For example, when the number of classes $k$ is $10$, $n$ is $256$. When the network training is over, due to the effect of loss function, the features of the training samples are basically distributed in the $k-1$ dimensions subspace \cite{9444709}. From this point of view, it seems that the remaining $n-k+1$ dimensions are useless, but the fact is that if $n$ is set to $k-1$ at the beginning, the classification accuracy drops significantly, which indicates that these extra dimensions play a role in optimizing classification features during the training process.

In this case, in order to further improve the utilization of all latent features, constraining the correlation between the dimensions is proposed. For example, for the image classification of cats and mice, if the correlation between the dimensions is not restricted, one of the learned features may represent the body size and the other dimension represents the facial contour size. Obviously, these two features have a strong correlation, which means a large body must have a large facial contour. For classification, the classification performance based on the two features is very similar or even the same as that based on one of the two features. Therefore, the resources occupied by one of the dimensions are wasted. When the constraint of the correlation between the dimensions is added, the two features learned in the above example may be hair color and body size, which are almost irrelevant. Compared with the classification based on a single feature, it is obvious that the combination of the two is better for classification and achieves the purpose of making full use of features.

Barlow Twins Loss adds a decorrelation mechanism to reduce the redundancy between network output features, so that the output features contain non-redundant information of samples. Barlow Twins uses a cross-correlation matrix and punishes the non-diagonal terms of the calculated cross-correlation matrix. Our method draws on this decorrelation mechanism, but instead of using the cross-correlation matrix, the self-correlation matrix of the difference between the latent features of the samples and the predefined central features is used. The non-diagonal terms are also punished to constrain the correlation between the different dimensions of the latent features of samples. Therefore, Self-correlation Constraint Loss is proposed:
\begin{normalsize}
\begin{equation}
L_{SC}=\sum_{i}^n{\sum_{j \neq i}^n{R_{ij}^2}}
\end{equation}
\end{normalsize}
\begin{normalsize}
\begin{equation}
R=\frac{1}{B-1}{(X-X_{pedcc})(X-X_{pedcc})^T}
\end{equation}
\end{normalsize}
where $n$ represents feature dimension, $R$ represents the self-correlation matrix of the difference matrix formed by the difference between the sample features and the predefined central features. The element in the $i$-th row and $j$-th column of the self-correlation matrix is the correlation coefficient between the $i$-th column and the $j$-th column of the difference matrix. $B$ is the number of samples in each batch. $X$ is the normalized feature matrix of $B$ samples in Fig. 1, where each column vector represents a sample. $X_{pedcc}$ is the PEDCC matrix corresponding to the sample feature.

\subsection{POD Loss}
On the one hand, Norm-adaptive Cosine (NaC) Loss is used to constrain the distance between the sample feature and its class PEDCC feature center, and on the other hand it accelerates the convergence speed of network training. At the same time, Self-correlation Constraint (SC) Loss is defined as the constraint on the correlation between the different dimensions of sample latent features, that is, the decorrelation term, which includes punishing the off-diagonal terms of the self-correlation matrix of the difference between the sample features and the predefined central features. The initial expression of POD Loss is:
\begin{normalsize}
\begin{equation}
L_{POD}=L_{NaC}+\lambda{L_{SC}}
\end{equation}
\end{normalsize}
where $L_{NaC}$ is the norm-adaptive Cosine loss function, $L_{SC}$ is the loss function that restricts the correlation between feature dimensions, and $\lambda$ is the weighting coefficient. For some experiments, the adjustment of $\lambda$ may improve the classification performance.
    
\section{Experiments and results}
Experiment is implemented using Pytorch1.0 \cite{pytorch}. The network structure used in the whole experiment defaults to ResNet50 \cite{7780459} (unless specified in detail). The datasets used include CIFAR10 \cite{Krizhevsky2009LearningML}, CIFAR100 \cite{Krizhevsky2009LearningML}, Tiny ImageNet \cite{5206848}, FaceScrub \cite{7025068} and ImageNet \cite{5206848} dataset. In order to make the network structure more suitable for the image sizes of different datasets, some modifications are made to the original ResNet50 structure. For CIFAR10 (images size is $32 \times 32$), CIFAR100 (images size is $32 \times 32$) and Tiny ImageNet (images size is $64 \times 64$), the convolution kernel of the first convolution layer is changed from the original $7 \times 7$ to $3 \times 3$, and the step size is changed to $1$. The maximum pooling layer in the second convolutional layer is also eliminated (except Tiny ImageNet). The above changes are not made for ImageNet. In addition, the number of feature dimensions input by the PEDCC layer in the network structure varies with the number of classes of these datasets. All experimental results are the average of three experiments.

\subsection{Experimental datasets}
CIFAR10 dataset contains $10$ classes of RGB color images, and the images size is $32 \times 32$. There are $50,000$ training images and $10,000$ test images in the dataset. CIFAR100 dataset contains $100$ classes of images, the images size is $32 \times 32$, a total of $50,000$ training images and $10,000$ test images. FaceScrub dataset contains $100$ classes of images, the images size is $64 \times 64$, there are $15896$ training images and 3896 test images in the dataset. Tiny ImageNet dataset contains $200$ classes of images, the images size are $64 \times 64$, there are a total of $100,000$ training images and $10,000$ test images. For these datasets, standard data enhancement \cite{10.1145/3065386} is performed, that is, the training images are filled with $4$ pixels, randomly cropped to the original size, and horizontally flipped with a probability of $0.5$, and the test images are not processed.

ImageNet dataset contains $1000$ classes of images. The images size is not unique, and the height and width are both greater than $224$. There are $1282166$ training images and $51000$ test images in the dataset. For this dataset, the training images are randomly cropped to different sizes and aspect ratios, scaled to $224 \times 224$, and flipped horizontally with a probability of $0.5$. The test images are scaled to $256$ proportionally with a small side length, and then the center is cropped to $224 \times 224$.

For the above datasets, in the training phase, using the SGD optimizer, the weight decay is $0.0005$ (ImageNet is $0.0001$), and the momentum is $0.9$. The initial learning rate is $0.1$, and a total of $100$ epochs are trained. At the $30th$, $60th$, and $90th$ epoch, the learning rate drops to one-tenth of the original, and the batchsize is set to $128$ (ImageNet is $96$).

\subsection{Experimental results}
\subsubsection{Selection of feature dimensions}
According to the theoretical analysis of feature dimensions in Section 3, comparison experiments are carried out on the selection of multiple different feature dimensions for datasets of different classes. $\alpha$ and $\lambda$, the parameters in POD Loss, are set to 0.01 and 1. TABLE 2 shows the classification performance under different feature dimensions on CIFAR10 dataset. Within a certain range, increasing the feature dimension will get better classification performance, but when the number of features reaches a certain scale, the performance of the classifier is declining. The feature dimensions with the best performance on different datasets under multiple experiments are shown in TABLE 3. For CIFAR10 dataset, $10$ evenly-distributed class centroids are predefined in a $256$ dimensions hyperspace. For CIFAR100, FaceScrub and Tiny ImagNet dataset, several evenly-distributed class centroids of corresponding classes are respectively predefined in a $512$ dimensions hyperspace. For ImageNet dataset, the top $30$ classes, the top $100$ classes, the top $200$ classes, the top $500$ classes, and all the $1000$ classes are selected for experiments. The optimal feature dimensions are $256$, $512$, $512$, $1024$ and $2048$.

\begin{table}[t]
\caption{Classification accuracy (\%) under different feature dimensions on CIFAR10 dataset}
\centering
\setlength{\tabcolsep}{12mm}{
\begin{tabular}{lc}
\toprule
  Feature Dimension & Accuracy(\%) 
  \\
\midrule
  9 & 92.71 
  \\
  32 & 93.51
  \\
  64 & 93.77
  \\
  128 & 93.98
  \\
  256 & \textbf{94.31} 
  \\
  512 & 94.25
  \\
  1024 & 93.82
  \\
\bottomrule
\end{tabular}}
\end{table}

\begin{table}[t]
\caption{The number of classes and the optimal feature dimensions of different datasets}
\centering
\setlength{\tabcolsep}{3.5mm}{
\begin{tabular}{lcc}
\toprule
  Dataset & Number of Classes & Feature Dimension 
  \\
\midrule
  CIFAR10 & 10 &  256 
  \\
  CIFAR100 & 100 & 512 
  \\
  FaceScrub & 100 & 512
  \\
  Tiny ImageNet & 200 & 512
  \\
  ImageNet(30) & 30 & 256 
  \\
  ImageNet(100) & 100 & 512 
  \\
  ImageNet(200) & 200 & 512 
  \\
  ImageNet(500) & 500 & 1024 
  \\
  ImageNet(1000) & 1000 & 2048 
  \\
\bottomrule
\end{tabular}}
\begin{tablenotes}
\footnotesize
\item[1] ImageNet($N$): The top $N$ classes of ImageNet dataset.
\end{tablenotes}
\end{table}

\subsubsection{Role of Norm-adaptive Cosine Loss}
Comparative experiments on Softmax, Center Loss, AM-Softmax$(m=0.35)$, ArcFace$(m=0.5)$\cite{8953658}, CM$(m1=0.9, m2=0.4, m3=0.15)$\cite{8953658} \cite{app11146545}, COT, PEDCC-Loss$(m=0.5)$ , Cosine Loss and NaC Loss$(\alpha=0.01)$ are conducted on some small-size datasets. The experimental results are shown in TABLE 4. For face recognition task (FaceScrub), Center Loss,  AM-Softmax$(m=0.35)$, ArcFace$(m=0.5)$ and CM$(m1=0.9, m2=0.4, m3=0.15)$ have better classification performance than Softmax Loss. But for general image classification task (CIFAR10, CIFAR100 and Tiny ImageNet), their classification performance is not always better than Softmax. Although COT and PEDCC-Loss have improved the classification performance of these two tasks, the improvement is limited. Compared with Cosine Loss, the classification accuracy of NaC loss has been greatly improved in these datasets, which obviously shows the importance of the adaptive additional value $\delta(e,M)$ in our loss.
Among these loss functions, NaC Loss always has the highest classification performance on these datasets. In the network training process, the convergence speed of CM, COT, PEDCC-Loss and NaC Loss is shown in Fig. 4. As can be seen from Fig. 4, NaC Loss training has a faster and more stable convergence speed.

\begin{table*}[t]
\caption{Comparison of classification accuracy (\%) of various loss functions on small-size datasets}
\centering
\setlength{\tabcolsep}{6mm}{
\begin{tabular}{lcccc}
\toprule
  \diagbox{Loss}{Dataset} & CIFAR10 & CIFAR100 & Tiny ImageNet & FaceScrub 
  \\
\midrule
  Softmax Loss & 93.25 & 76.42 & 62.10 & 91.04 
  \\
  Center Loss & 92.82  (0.43$\downarrow$) & 76.91 (0.49$\uparrow$) & 60.71  (1.39$\downarrow$) & 92.82 (1.78$\uparrow$)
  \\
  AM-Softmax$(m=0.35)$ & 92.49  (0.76$\downarrow$) & 75.99 (0.43$\downarrow$) & 59.69  (2.41$\downarrow$) & 93.60 (2.56$\uparrow$)
  \\
  ArcFace$(m=0.5)$ & 92.44  (0.81$\downarrow$) & 76.41 (0.01$\downarrow$) & 62.15 (0.05$\uparrow$) & 93.93 (2.89$\uparrow$)
  \\
  CM$(m1=0.9, m2=0.4, m3=0.15)$ & 93.19  (0.06$\downarrow$) & 75.78 (0.64$\downarrow$) & 59.53 (2.57$\downarrow$) & 94.39 (3.35$\uparrow$)
  \\
  COT & 93.60  (0.35$\uparrow$) & 76.96 (0.54$\uparrow$) & 62.21 (0.11$\uparrow$) & 94.30 (2.26$\uparrow$)
  \\
  PEDCC-Loss$(m=0.5)$ & 93.65 (0.40$\uparrow$) & 77.04 (0.62$\uparrow$) & 62.97 (0.87$\uparrow$) & 94.20 (3.16$\uparrow$)
  \\
  Cosine Loss & 93.83 (0.58$\uparrow$) & 77.70  (1.28$\uparrow$) & 62.61  (0.51$\uparrow$) & 93.01 (1.97$\uparrow$) 
  \\
  NaC Loss$(\alpha=0.01)$ & \textbf{94.23 (0.98$\uparrow$)} & \textbf{78.24  (1.82$\uparrow$)} & \textbf{64.35  (2.25$\uparrow$)} & \textbf{95.10 (4.06$\uparrow$)} 
  \\
  POD Loss$(\alpha=0.01, \lambda=1)$ & \textbf{94.31 (1.06$\uparrow$)} & \textbf{79.17  (2.75$\uparrow$)} & \textbf{64.58  (2.48$\uparrow$)} & \textbf{95.22 (4.18$\uparrow$)} 
  \\
\bottomrule
\end{tabular}}
\begin{tablenotes}
\footnotesize
\item[1] CM$(m1=0.9, m2=0.4, m3=0.15)$\cite{8953658} \cite{app11146545}: Combined Margin Loss, multiplicative angular margin $m1$, additive angular margin $m2$, and additive Cosine margin $m3$.
\end{tablenotes}
\end{table*}

\begin{figure}
  \centering
  \includegraphics[width=0.48\textwidth]{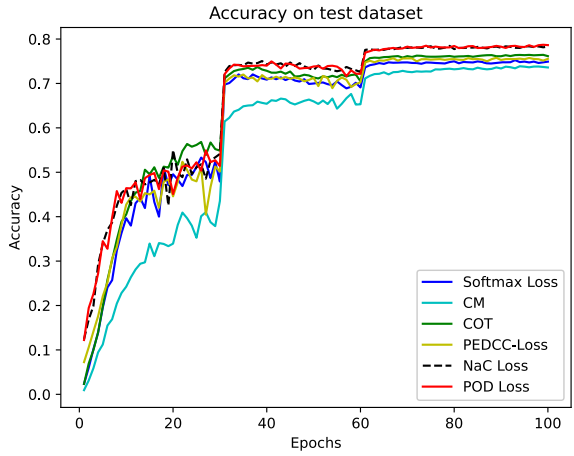}
\caption{Variation of classification accuracy of various loss functions with epoch during training.}
\end{figure}


\subsubsection{Role of SC Loss}
$\alpha$ and $\lambda$, the parameters in POD Loss, are set to 0.01 and 1. The variance of the $2048$ dimensions feature eigen-vectors (normalized) before the fully connected layer of the network is respectively calculated after POD Loss and NaC Loss training, as shown in TABLE 5. The variance of eigen-vectors after POD Loss training is smaller than that of NaC Loss, indicating that the existence of SC Loss can balance the energy of the features and gain some benefits for subsequent classification.

\begin{table}[t]
\caption{Variance of feature eigen-vectors of NaC Loss and POD Loss}
\centering
\setlength{\tabcolsep}{7mm}{
\begin{tabular}{lc}
\toprule
  Loss Function & Variance of Feature Eigen-vectors 
  \\
\midrule
  NaC Loss & 1.82e-8
  \\
  POD Loss & \textbf{1.53e-8}
  \\
\bottomrule
\end{tabular}}
\end{table}

Comparative experiments of NaC Loss and POD Loss have also been carried out on some small-size datasets, and the experimental results are shown in TABLE 4. The experimental results show that POD Loss with SC Loss is better than a single NaC Loss in classification accuracy on multiple datasets.


\subsubsection{POD Loss}
For parameter $\lambda$ in formula 16, the tendency chart of the value of total loss and $\lambda$ on four small-size datasets is shown in Fig. 5. As can be seen from Fig. 5, when $lg(\lambda)$ increases from $-2$ to $3$, the value of loss remains stable, and with the continuous increase, the loss value increases significantly. So we choose $lg(\lambda)=0$ (i.e. $\lambda=1$) to complete all other experiments.
\begin{figure}
  \centering
  \includegraphics[width=0.48\textwidth]{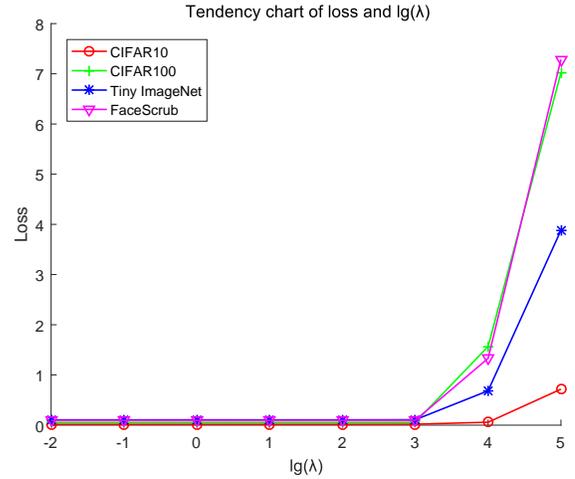}
\caption{Tendency chart of loss and $\lambda$.}
\end{figure}

Comparative experiments of Softmax Loss and POD Loss are conducted on the above datasets. $\alpha$ and $\lambda$, the parameters in POD Loss, are set to 0.01 and 1. The experimental results are shown in TABLE 6, and the values in parentheses show how much the accuracy of POD Loss is improved compared with that of Softmax Loss. Experimental results show that the classification performance of POD Loss is higher than that of Softmax Loss on multiple datasets, and the classification accuracy is much higher than Softmax Loss on some datasets. Experiments on the ImageNet dataset show that with the increase of the number of classes in the dataset, the classification accuracy of POD Loss is gradually approaching Softmax Loss. The reason lies in the structure of the network itself (the maximum output feature dimension of ResNet50 is $2048$, the ratio of the feature dimension to the number of large classes is much smaller than the ratio of the feature dimension to the number of small classes, and there are fewer or no extra dimensions to benefit classification). So, on large-class datasets, the advantages of POD Loss over Softmax Loss are limited.

\begin{table}[t]
\caption{Comparison of classification accuracy (\%) of Softmax Loss and POD Loss on different datasets}
\centering
\setlength{\tabcolsep}{4.2mm}{
\begin{tabular}{lcc}
\toprule
  \diagbox{Dataset}{Loss\\ Function} & Softmax Loss & POD Loss 
  \\
\midrule
  CIFAR10 & 93.25 &  \textbf{94.31 (1.06$\uparrow$)} 
  \\
  CIFAR100 & 76.42 & \textbf{79.17 (2.75$\uparrow$)} 
  \\
  Tiny ImageNet & 62.10 & \textbf{64.58 (2.48$\uparrow$)}
  \\
  FaceScrub & 91.04 & \textbf{95.22 (4.18$\uparrow$)}
  \\
  ImageNet(30) & 79.86 & \textbf{86.47 (6.61$\uparrow$)}
  \\
  ImageNet(100) & 78.25 & \textbf{82.36 (4.11$\uparrow$)}
  \\
  ImageNet(200) & 82.06 & \textbf{83.74 (1.68$\uparrow$)}
  \\
  ImageNet(500) & 82.14 & \textbf{82.56 (0.42$\uparrow$)}
  \\
  ImageNet(1000) & 75.65 & \textbf{76.08 (0.43$\uparrow$)}
  \\
\bottomrule
\end{tabular}}
\end{table}

At the same time, in the process of network training, the convergence speed of POD Loss and Softmax Loss is shown in Fig. 4. As can be seen from Fig. 4, POD Loss training converges faster.

The visualization of feature $\bm{x}$ after FC1 on two-dimensional space is performed, and compared with other various loss functions, as shown in Fig. 6. For making the performance of visualization better, the first five classes of MNIST \cite{6296535} dataset is chosen for the experiment. CM represents the combined margin loss, including multiplicative angular margin, additive angular margin, and additive Cosine margin. As can be seen from Fig. 6, in the Euclidean space, compared with Softmax, the inter-class intervals of Center Loss, AM-Softmax and CM are larger, but the included angles between centers of different classes are not even, which means the distance between similar classes is closer, and the number of samples close to the origin is increased, which are easy to be classified incorrectly. The introduction of complement entropy loss in COT makes the class center distribution more uniform and the intra-class distance more compact, but the improvement is obviously limited. Although PEDCC-Loss introduces PEDCC to improve the disadvantage of uneven angle between class centers, there are still many samples near the base point, which has a bad impact on the classification accuracy. POD Loss ($\alpha = 0, \lambda = 1$) overcomes the shortcomings of PEDCC-Loss, but the intra-class distance is too large, resulting in overlapping parts. On this basis, the weighted coefficient $\alpha$ in the adaptive additional value $\delta(e,M)$ is gradually increased. It can be seen from subfigure (g) that the inter-class distance becomes larger and more uniform, which means that the classification boundary is clearer, the classification result is more accurate and and the robustness of network is higher. However, the weighted coefficient $\alpha$ is not the bigger the better, when $\alpha$ increases to $0.1$, the feature distribution of the sample is confused and the classification performance is seriously reduced. Therefore, $\alpha=0.01$ is selected to complete the comparative experiment in this article, which shows the optimal feature distribution.

\begin{figure*}
	\centering
	\begin{subfigure}{0.24\textwidth}
		\centering
		\includegraphics[width=4.5cm,height=4.5cm]{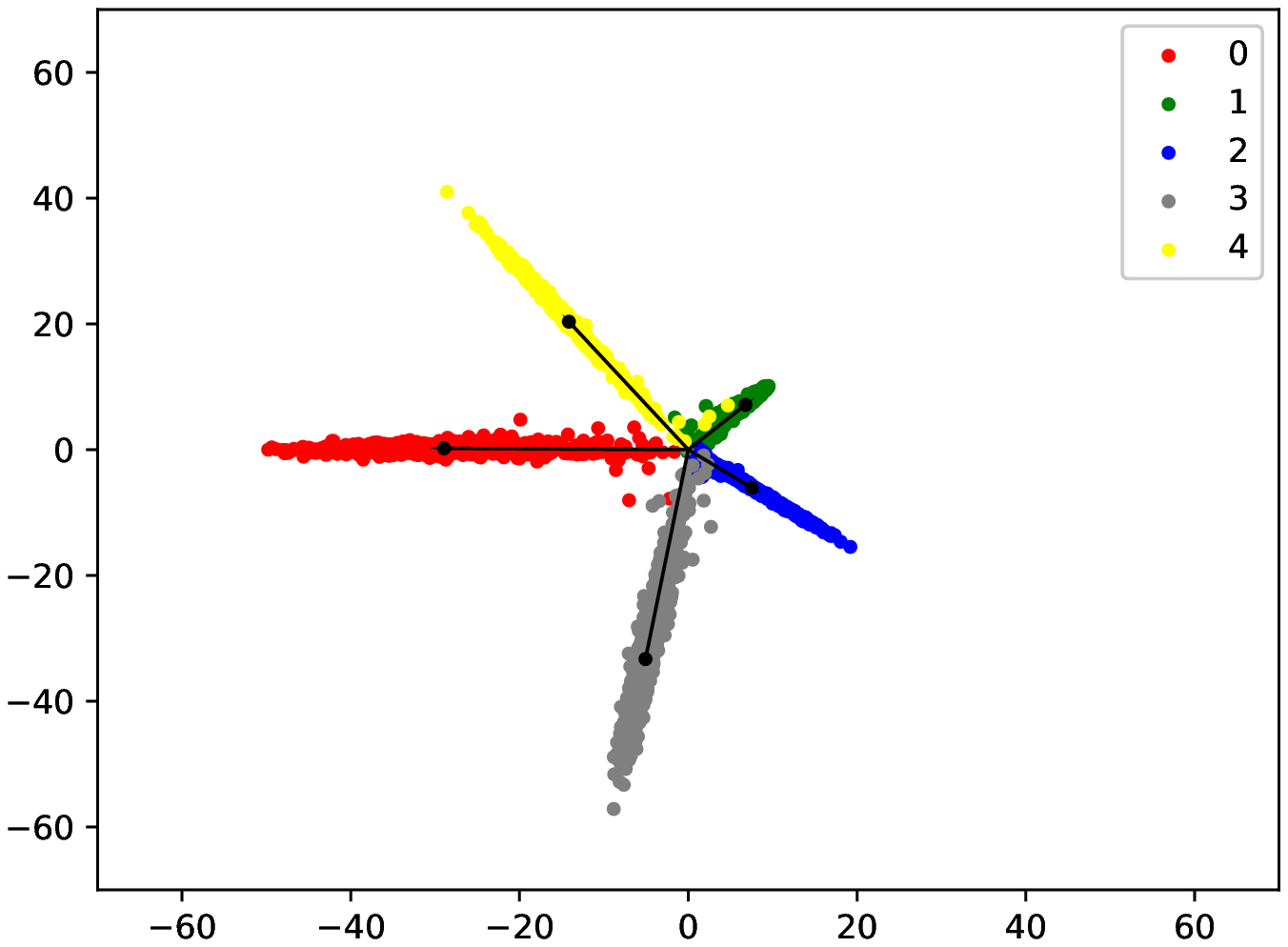}
		\caption{Center Loss}
	\end{subfigure}
	\centering
	\begin{subfigure}{0.24\textwidth}
		\centering
		\includegraphics[width=4.5cm,height=4.5cm]{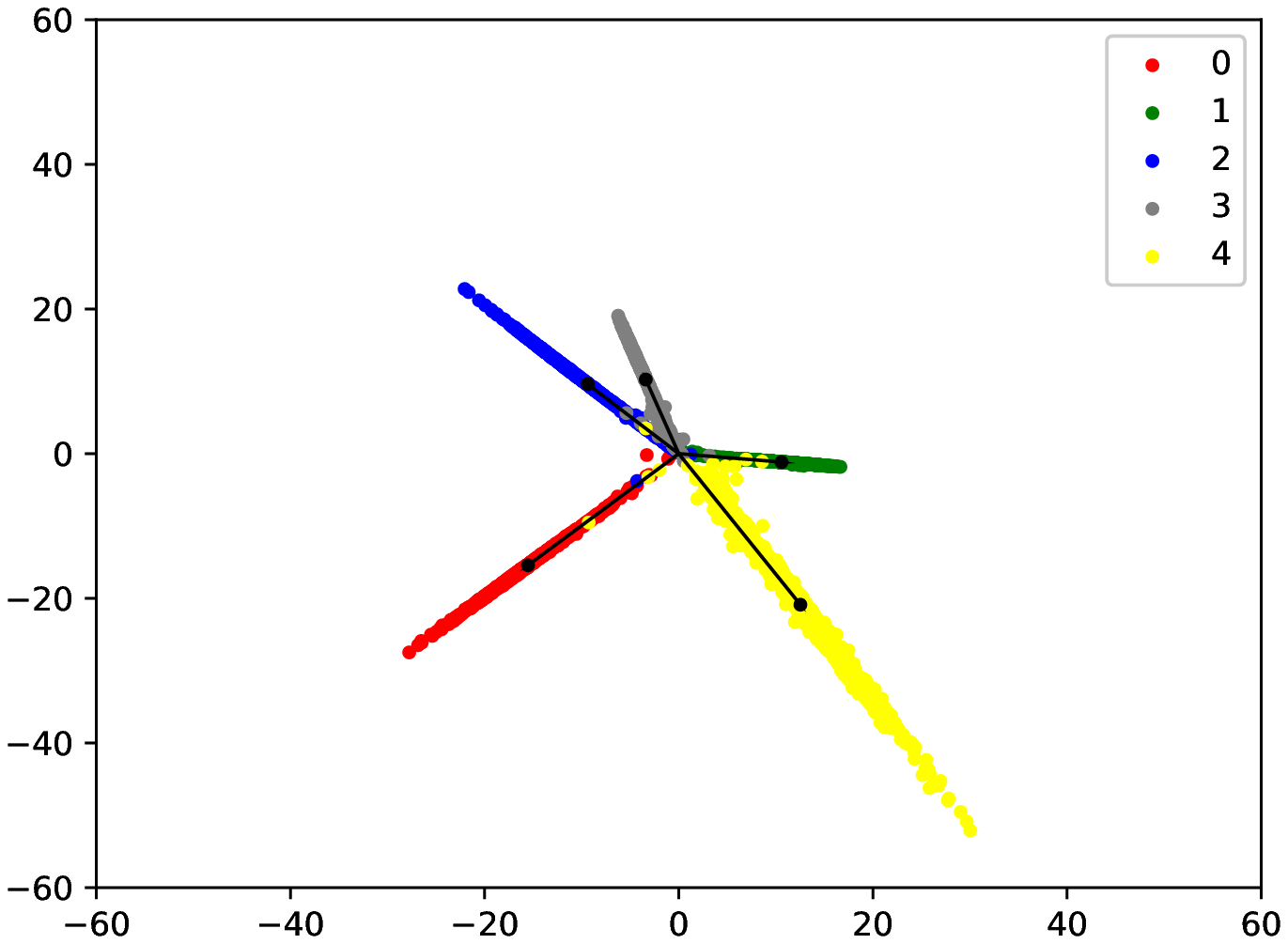}
		\caption{AM-Softmax}
	\end{subfigure}
	\centering
	\begin{subfigure}{0.24\textwidth}
		\centering
		\includegraphics[width=4.5cm,height=4.5cm]{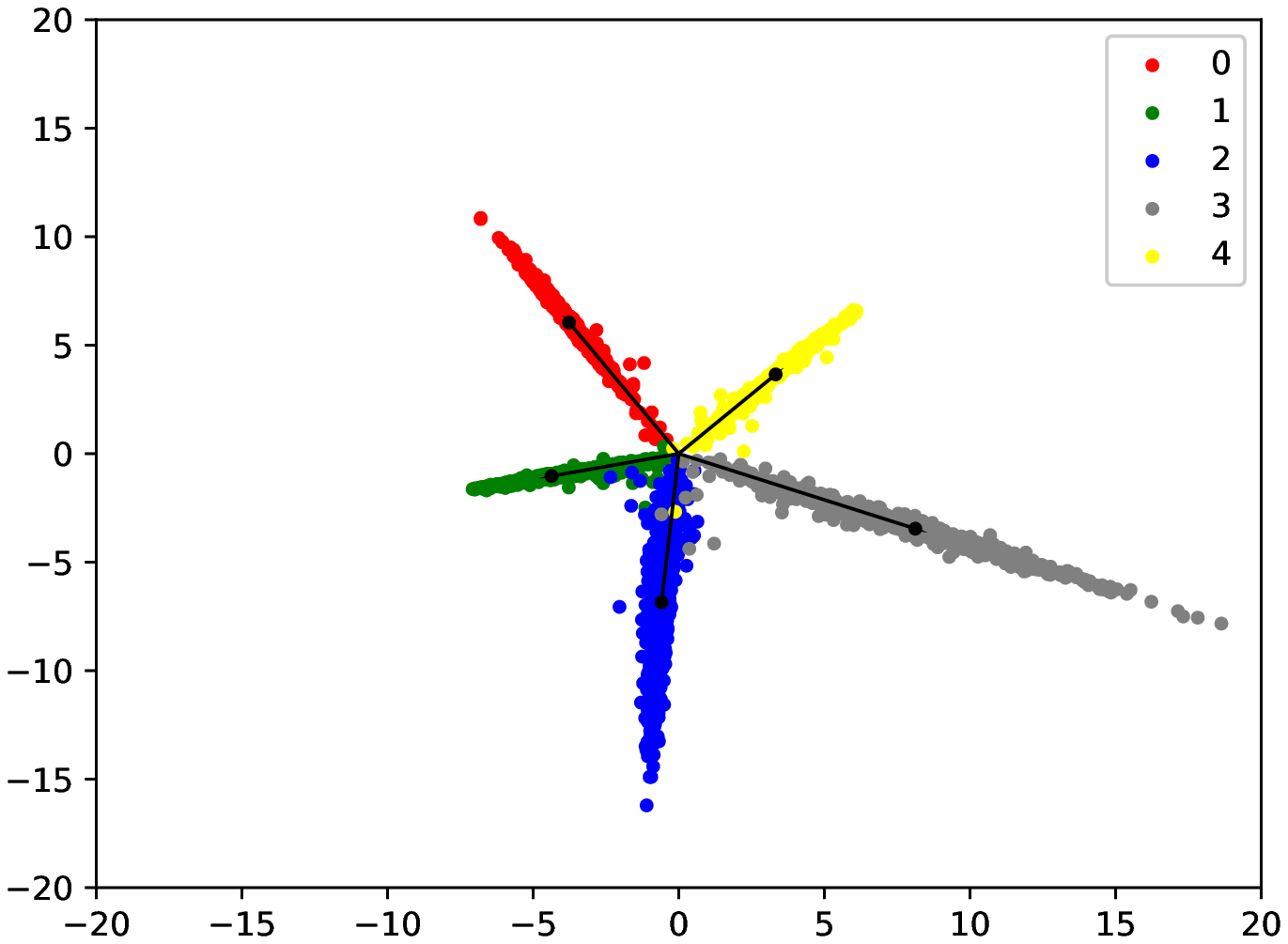}
		\caption{CM}
	\end{subfigure}
	\begin{subfigure}{0.24\textwidth}
	    \centering
		\includegraphics[width=4.5cm,height=4.5cm]{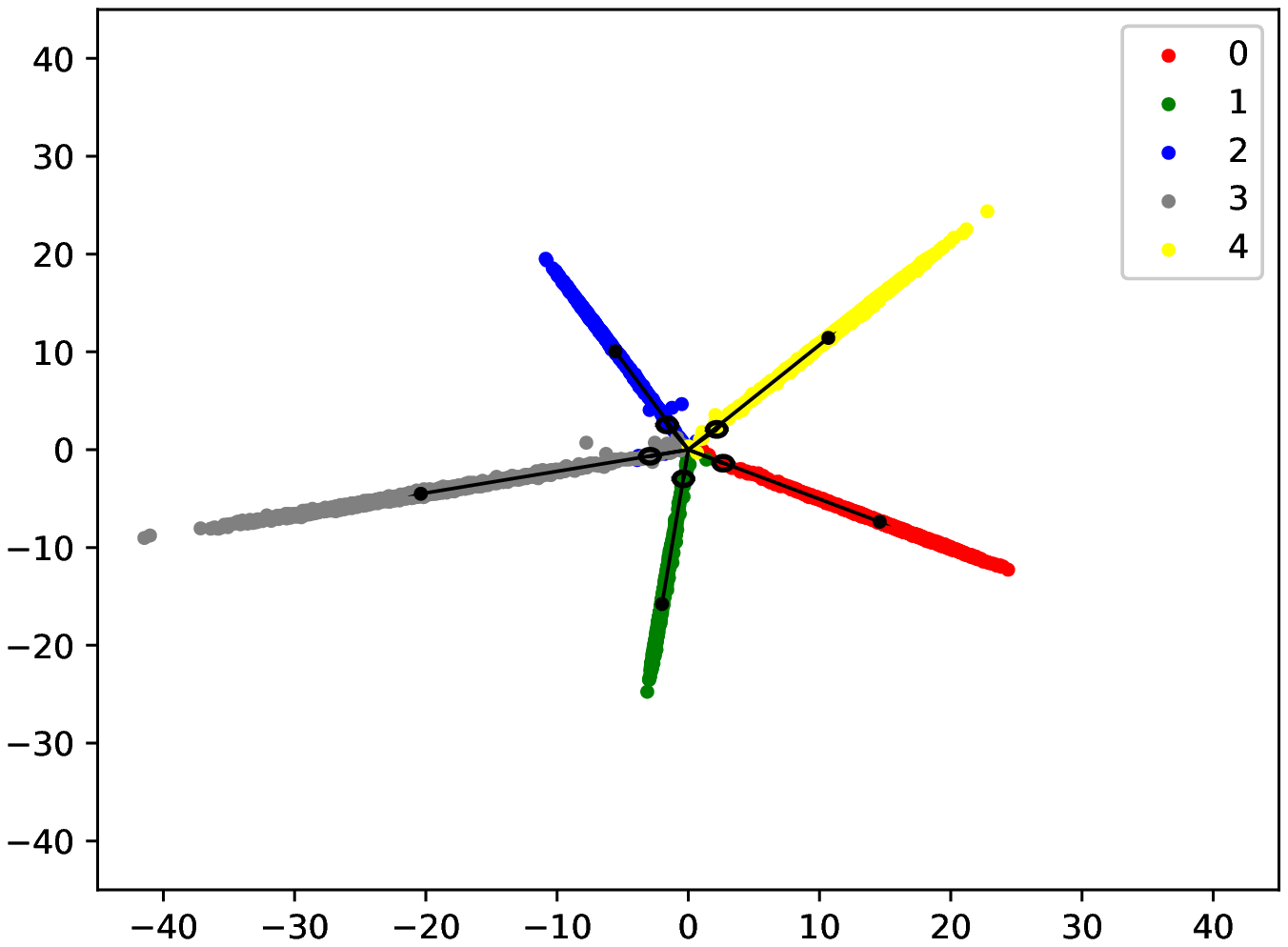}
		\caption{PEDCC-Loss}
	\end{subfigure}
	\centering
	\begin{subfigure}{0.24\textwidth}
		\centering
		\includegraphics[width=4.5cm,height=4.5cm]{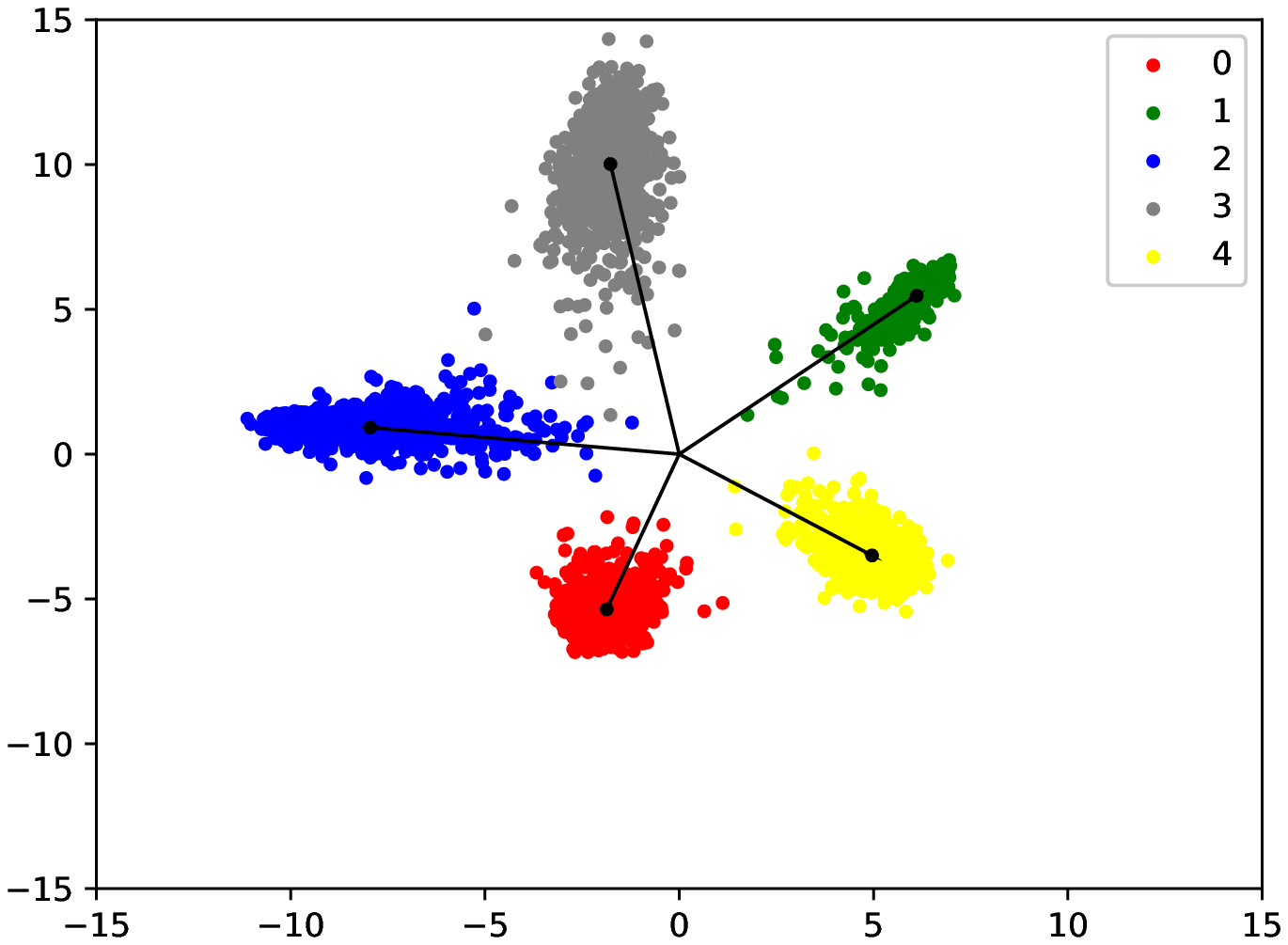}
		\caption{Softmax}
	\end{subfigure}
	\begin{subfigure}{0.24\textwidth}
	    \centering
		\includegraphics[width=4.5cm,height=4.5cm]{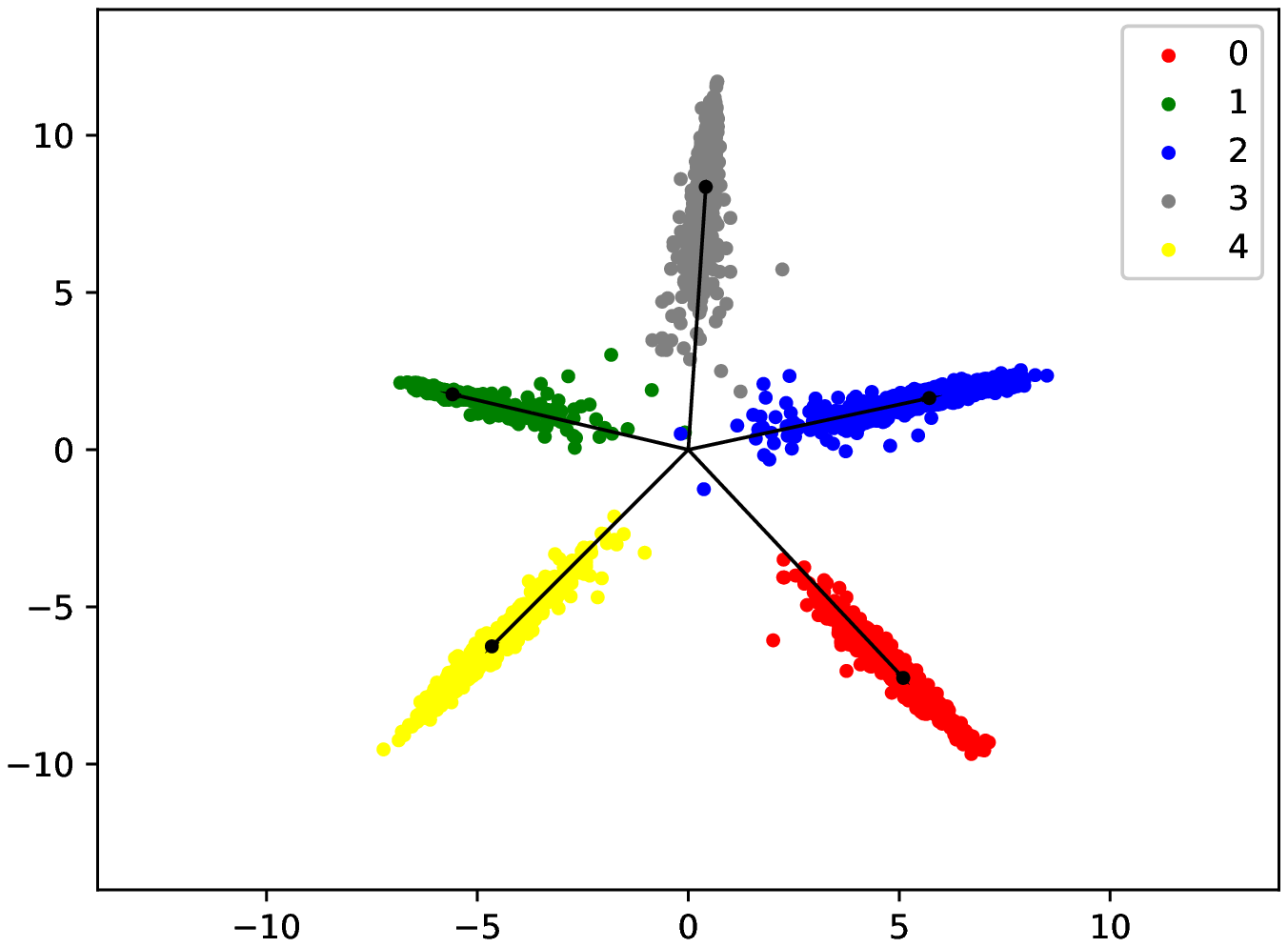}
		\caption{COT}
	\end{subfigure}
	\centering
	\begin{subfigure}{0.24\textwidth}
		\centering
		\includegraphics[width=4.5cm,height=4.5cm]{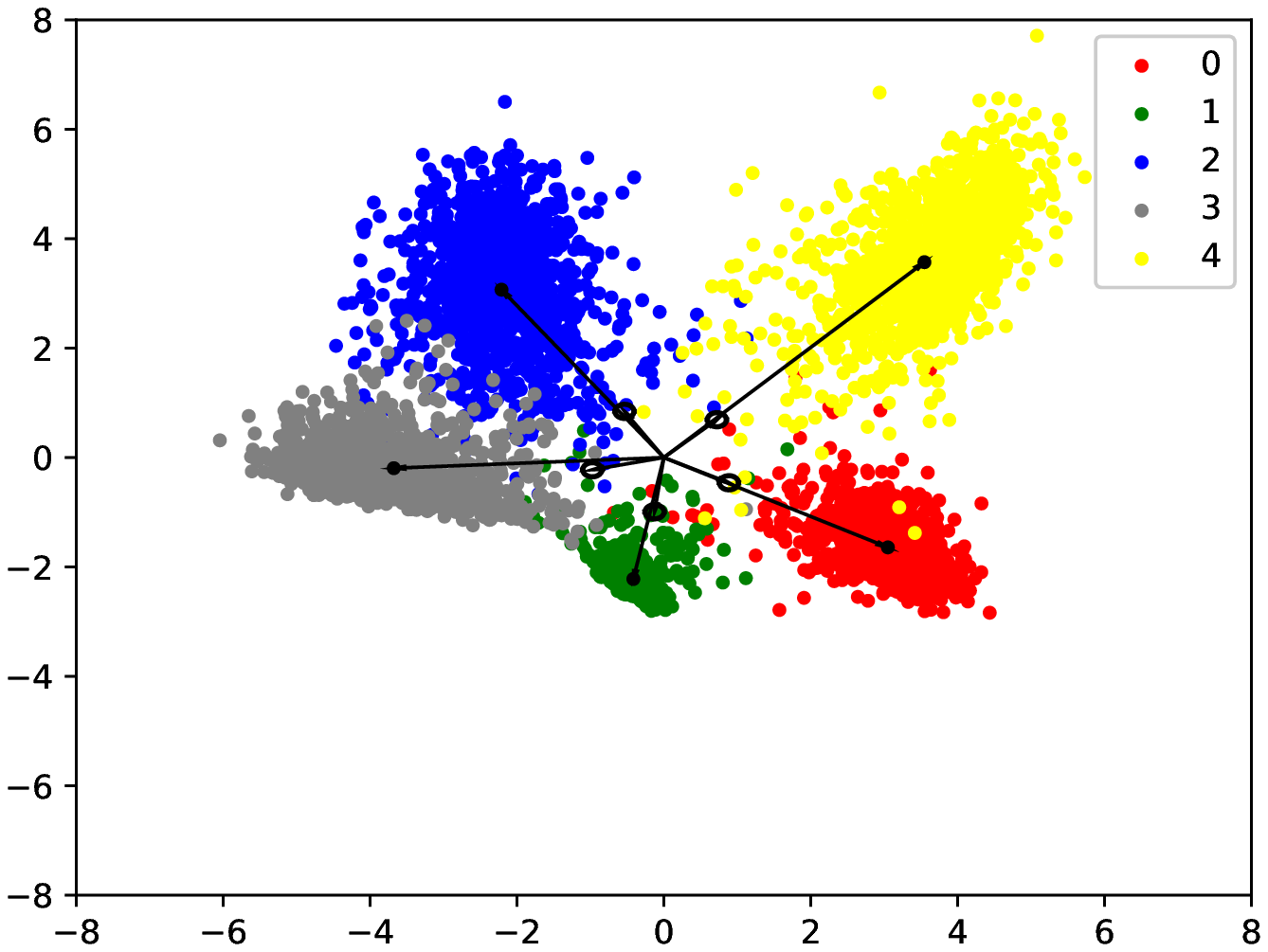}
		\caption{POD Loss ($\alpha=0,\lambda=1$)}
	\end{subfigure}
	\centering
	\begin{subfigure}{0.24\textwidth}
		\centering
		\includegraphics[width=4.5cm,height=4.5cm]{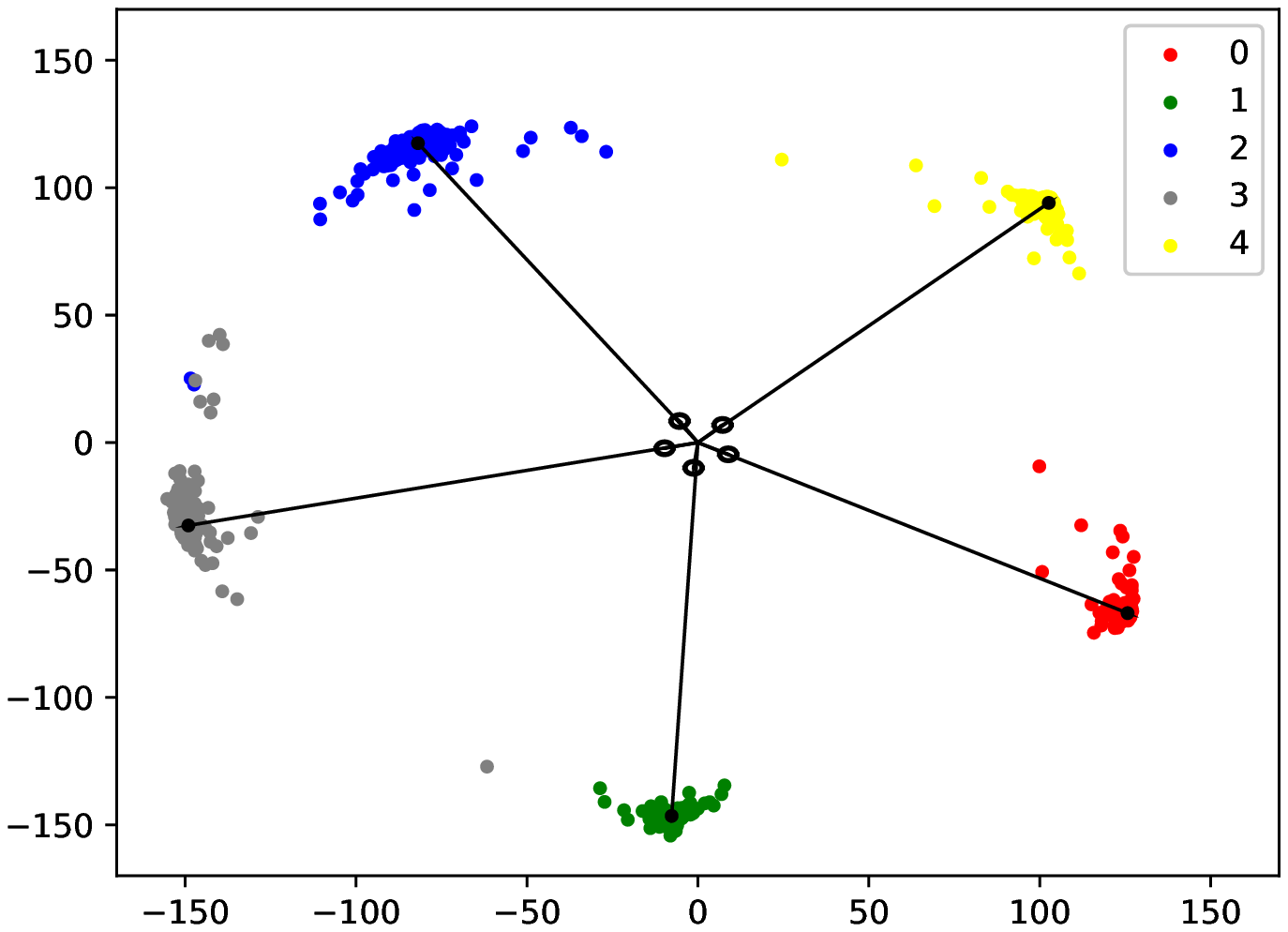}	
		\caption{POD Loss ($\alpha=0.01,\lambda=1$)}
	\end{subfigure}
	\caption{Visualization of feature $\bm{x}$ after FC1 for various loss functions on 2-D space, and POD Loss ($\alpha=0.01,\lambda=1$) has optimal feature distribution. Each color represents a class and the black spots in the center of each class is the mean point of the sample. In (d), (g) and (h), the five black circles near the origin are predefined PEDCC points.}
\end{figure*}


\begin{table}[t]
\caption{Comparison of classification accuracy(\%) of Softmax Loss and POD Loss on different traditional networks}
\centering
\small
\setlength{\tabcolsep}{1.0mm}{
\begin{tabular}{lccc}
\toprule
 \tabincell{c}{Network} & Dataset & \tabincell{c}{Accuracy\\(Softmax Loss)} & \tabincell{c}{Accuracy\\(POD Loss)}
  \\
\midrule
\multirow{4}*{ResNet50} & CIFAR10 & 93.25 & \textbf{94.31 (1.06$\uparrow$)} \\
~ & CIFAR100 & 76.42 & \textbf{79.17  (2.75$\uparrow$)} \\
~ & Tiny ImageNet &62.10 & \textbf{64.58  (2.48$\uparrow$)} \\
~ & FaceScrub & 91.04 & \textbf{95.22  (4.18$\uparrow$)} \\
\midrule
  \multirow{4}*{DenseNet121} & CIFAR10 & 94.24 &  \textbf{94.61   (0.37$\uparrow$)} 
  \\
  ~ & CIFAR100 & 77.90 & \textbf{79.38   (1.48$\uparrow$)} 
  \\
  ~ & Tiny ImageNet & 61.76 & \textbf{64.12  (2.36$\uparrow$)} 
  \\
  ~ & FaceScrub & 95.29 & \textbf{95.81  (0.52$\uparrow$)} 
  \\
\midrule
  \multirow{4}*{MobileNetV2} & CIFAR10 & 84.31 &  \textbf{86.00  (1.69$\uparrow$)} 
  \\
  ~ & CIFAR100 & 52.41 & \textbf{55.66  (3.25$\uparrow$)} 
  \\
  ~ & Tiny ImageNet & 46.34 & \textbf{49.90  (3.56$\uparrow$)} 
  \\
  ~ & FaceScrub & 87.84 & \textbf{88.88 (1.04$\uparrow$)} 
  \\
\bottomrule
\end{tabular}}
\end{table}

\subsubsection{POD Loss on different network structures}
Comparative experiments of Softmax Loss and POD Loss are conducted on ResNet50, DenseNet121 \cite{2016Densely} and MobileNetV2 \cite{8578572}. $\alpha$ and $\lambda$, the parameters in POD Loss, are set to 0.01 and 1. The experimental results are shown in TABLE 7, and the values in parentheses show how much the accuracy of POD Loss is improved compared with that of Softmax Loss. Experimental results show that the classification performance of POD Loss is higher than that of Softmax Loss on these three typical CNN network structures.

The tremendous success of the Transformer in language and other fields has led researchers to investigate its adaptation to computer vision, where it has demonstrated promising results on image classification tasks. Comparative experiments of Softmax Loss and POD Loss are conducted on ViT-B/16 \cite{2021An} and Swin-B \cite{2021arXiv210314030L}. $\alpha$ and $\lambda$, the parameters in POD Loss, are set to 0.01 and 1. In order to highlight the advantages of transformer networks, the images size of all datasets in this experiment is resized to 224 and the training settings are different from the above. For ViT-B/16, the training setting mostly follows ViT \cite{2021An}, that is, loading the parameters pre-trained on the ImageNet-21k dataset , fine-tuning the model using the SGD optimizer with a momentum of 0.9 and a weight decay of 0, and using a linear learning rate warmup. For Swin-B, the training setting mostly follows Swin Transformer \cite{2021arXiv210314030L}, that is, employing an AdamW \cite{2014Adam} optimizer for 300 epochs using a Cosine decay learning rate scheduler and 20 epochs of linear warm-up. A batch size of 256, an initial learning rate of 0.001, and a weight decay of 0.05 are used. The results in TABLE 8 shows the classification performance of POD Loss is always better than that of Softmax Loss on popular transformer networks.

The results in TABLE 8 show that the classification performance of POD Loss is significantly better than that of Softmax Loss on the first four datasets on popular transformer networks. On ImageNet datasets with 1000 classes, the recognition accuracy is slightly lower than that of Softmax Loss. We think there are two reasons: First, the feature dimension $n$ after the last feature extraction layer of ViT-B/16 and Swin-B is not within the range of $2*k$ and $40*k$ ($k$ is the number of classes), and the feature dimension $n$ of ViT-B/16 is only $768$ ($k \leq n+1$ is not satisfied); Second, for datasets with too many classes, the optimal distribution in the high-dimensional space is more difficult to be satisfied by network learning. On the contrary, it may be easier to optimize for Softmax Loss that takes the maximum value. 

\begin{table}[t]
\caption{Comparison of classification accuracy(\%) of Softmax Loss and POD Loss on different transformer networks}
\centering
\small
\setlength{\tabcolsep}{1.8mm}{
\begin{tabular}{lccc}
\toprule
 \tabincell{c}{Network} & Dataset & \tabincell{c}{Accuracy\\(Softmax Loss)} & \tabincell{c}{Accuracy\\(POD Loss)}
  \\
\midrule
\multirow{5}*{ViT-B/16} & CIFAR100 & 92.14 & \textbf{92.84 (0.70$\uparrow$)} \\
~ & FaceScrub & 97.28 & \textbf{97.69 (0.41$\uparrow$)} \\
~ & ImageNet(100) & 86.66 & \textbf{87.48 (0.82$\uparrow$)} 
\\
~ & ImageNet(500) & 86.38 & \textbf{87.11 (0.73$\uparrow$)} 
\\
~ & ImageNet(1000) & \textbf{82.16} & 81.20 (0.94$\downarrow$) 
\\
\midrule
\multirow{5}*{Swin-B} & CIFAR100 & 91.67 & \textbf{92.26  (0.59$\uparrow$)} \\
~ & FaceScrub & 99.13 & \textbf{99.26  (0.13$\uparrow$)} \\
~ & ImageNet(100) & 89.02 & \textbf{89.14 (0.12$\uparrow$)} 
\\
~ & ImageNet(500) & 88.74 & \textbf{88.95 (0.21$\uparrow$)} 
\\
~ & ImageNet(1000) & \textbf{84.68} & 84.53 (0.15$\downarrow$) 
\\
\bottomrule
\end{tabular}}
\begin{tablenotes}
\footnotesize
\item[1] ImageNet($N$): The top $N$ classes of ImageNet dataset.
\end{tablenotes}
\end{table}

\subsubsection{Discussion on the last classification method}
The constraint on latent features only solves the optimization problem of latent feature distribution, and the pattern classification method needs to be matched later. On the one hand, the POD Loss proposed in this article includes NaC Loss that constrains the latent features and the Cosine distance between the PEDCC centroids. When POD Loss converges, the Cosine distance between the latent feature vector and the PEDCC centroid of the correct label reaches the minimum, and the Cosine distances between the latent feature vector and the PEDCC centroids of the incorrect labels reach the maximum in a uniform state. The classification method is:
\begin{normalsize}
\begin{equation}
I=\mathop{argmax}\limits_{i=1,2,...,k}{(\bm{x}\cdot \bm{\tilde{pedcc_i}})}
\end{equation}
\end{normalsize}
\begin{normalsize}
\begin{equation}
=\mathop{argmax}\limits_{i=1,2,...,k}{(||\bm{x}|| \cdot ||\bm{\tilde{pedcc_i}}|| \cdot cos\theta_{\bm{x},\bm{\tilde{pedcc_i}}})}
\end{equation}
\end{normalsize}
\begin{normalsize}
\begin{equation}
=\mathop{argmax}\limits_{i=1,2,...,k}{(||\bm{x}|| \cdot cos\theta_{\bm{x},\bm{\tilde{pedcc_i}}})}
\end{equation}
\end{normalsize}
\begin{normalsize}
\begin{equation}
=\mathop{argmax}\limits_{i=1,2,...,k}{(cos\theta_{\bm{x},\bm{\tilde{pedcc_i}}})}
\end{equation}
\end{normalsize}
where $k$ represents the number of classes, $\bm{x}$ is the sample feature after FC1, $\bm{\tilde{pedcc_i}}$ is the PEDCC center of the $i$-th class after normalized, and $cos\theta_{\bm{x},\bm{\tilde{pedcc_i}}}$ represents the Cosine value between $\bm{x}$ and $\bm{\tilde{pedcc_i}}$.

On the other hand, under the premise that the features of samples satisfy the gaussian distribution, the mean and covariance of each PEDCC are different, and each class has different class conditional probability densities, so Gaussian Discriminant Analysis (GDA) method can also be used for classification.

\begin{table}[t]
\caption{Comparison of classification accuracy (\%) of two classification methods}
\centering
\setlength{\tabcolsep}{1.0mm}{
\begin{tabular}{lcc}
\toprule
  \diagbox{Dataset}{Method} & Cosine distance & Gaussian discrimination analysis 
  \\
\midrule
  CIFAR10 & \textbf{94.31 (0.08$\uparrow$)} &  94.23 
  \\
  CIFAR100 & \textbf{79.17 (4.80$\uparrow$)} & 74.37 
  \\
  Tiny ImageNet & \textbf{64.58 (13.49$\uparrow$)} & 51.09
  \\
\bottomrule
\end{tabular}}
\end{table}

Therefore, this article conducts a comparative experiment on the two classification methods of Cosine distance and GDA based on POD Loss. The experimental results on multiple datasets are shown in TABLE 9. It can be seen from TABLE 9 that the classification method of Cosine distance is better than GDA on multiple datasets. Therefore, Cosine distance is adopted as the final classification method in this article.

\section{Conclusion}
For deep learning classifier, this article proposes a Softmax-free loss function (POD Loss) based on predefined optimal-distribution of latent features. The loss function discards the constraints on the posterior probability in the traditional loss function, and only constrains the output of feature extraction to realize the optimal distribution of latent features, i.e. maximization of inter-class distance and minimization of intra-class distance. Our method includes norm-adaptive Cosine distance between sample feature vector and predefined evenly-distributed class centroids, the decorrelation mechanism between sample features, and finally the classification through the solidified PEDCC layer. The experimental results show that, compared to the commonly used Softmax Loss and its improved Loss, POD Loss achieves better performance on image classification tasks, and is easier to train and converge. In the future, new loss function based on latent feature distribution optimization will be further studied from the perspective of distinguishing classification features and representation features, so as to further improve the recognition performance of the network.

\bibliographystyle{IEEEtran}
\bibliography{tcsvt_jrnl}

\begin{thebibliography}{10}
\providecommand{\url}[1]{#1}
\csname url@samestyle\endcsname
\providecommand{\newblock}{\relax}
\providecommand{\bibinfo}[2]{#2}
\providecommand{\BIBentrySTDinterwordspacing}{\spaceskip=0pt\relax}
\providecommand{\BIBentryALTinterwordstretchfactor}{4}
\providecommand{\BIBentryALTinterwordspacing}{\spaceskip=\fontdimen2\font plus
\BIBentryALTinterwordstretchfactor\fontdimen3\font minus
  \fontdimen4\font\relax}
\providecommand{\BIBforeignlanguage}[2]{{%
\expandafter\ifx\csname l@#1\endcsname\relax
\typeout{** WARNING: IEEEtran.bst: No hyphenation pattern has been}%
\typeout{** loaded for the language `#1'. Using the pattern for}%
\typeout{** the default language instead.}%
\else
\language=\csname l@#1\endcsname
\fi
#2}}
\providecommand{\BIBdecl}{\relax}
\BIBdecl

\bibitem{10.1145/3065386}
\BIBentryALTinterwordspacing
A.~Krizhevsky, I.~Sutskever, and G.~E. Hinton, ``Imagenet classification with
  deep convolutional neural networks,'' \emph{Commun. ACM}, vol.~60, no.~6, p.
  84–90, May 2017. [Online]. Available: \url{https://doi.org/10.1145/3065386}
\BIBentrySTDinterwordspacing

\bibitem{7298965}
J.~Long, E.~Shelhamer, and T.~Darrell, ``Fully convolutional networks for
  semantic segmentation,'' in \emph{2015 IEEE Conference on Computer Vision and
  Pattern Recognition (CVPR)}, 2015, pp. 3431--3440.

\bibitem{electronics9081188}
\BIBentryALTinterwordspacing
I.~Adjabi, A.~Ouahabi, A.~Benzaoui, and A.~Taleb-Ahmed, ``Past, present, and
  future of face recognition: A review,'' \emph{Electronics}, vol.~9, no.~8,
  2020. [Online]. Available: \url{https://www.mdpi.com/2079-9292/9/8/1188}
\BIBentrySTDinterwordspacing

\bibitem{7780834}
H.~Nam and B.~Han, ``Learning multi-domain convolutional neural networks for
  visual tracking,'' in \emph{2016 IEEE Conference on Computer Vision and
  Pattern Recognition (CVPR)}, 2016, pp. 4293--4302.

\bibitem{ZAFEIRIOU20151}
\BIBentryALTinterwordspacing
S.~Zafeiriou, C.~Zhang, and Z.~Zhang, ``A survey on face detection in the wild:
  Past, present and future,'' \emph{Computer Vision and Image Understanding},
  vol. 138, pp. 1--24, 2015. [Online]. Available:
  \url{https://www.sciencedirect.com/science/article/pii/S1077314215\\000727}
\BIBentrySTDinterwordspacing

\bibitem{Simonyan2015VeryDC}
K.~Simonyan and A.~Zisserman, ``Very deep convolutional networks for
  large-scale image recognition,'' \emph{CoRR}, vol. abs/1409.1556, 2015.

\bibitem{7780459}
K.~He, X.~Zhang, S.~Ren, and J.~Sun, ``Deep residual learning for image
  recognition,'' in \emph{2016 IEEE Conference on Computer Vision and Pattern
  Recognition (CVPR)}, 2016, pp. 770--778.

\bibitem{8099726}
G.~Huang, Z.~Liu, L.~Van Der~Maaten, and K.~Q. Weinberger, ``Densely connected
  convolutional networks,'' in \emph{2017 IEEE Conference on Computer Vision
  and Pattern Recognition (CVPR)}, 2017, pp. 2261--2269.

\bibitem{10.5555/3045118.3045167}
S.~Ioffe and C.~Szegedy, ``Batch normalization: Accelerating deep network
  training by reducing internal covariate shift,'' ser. ICML'15.\hskip 1em plus
  0.5em minus 0.4em\relax JMLR.org, 2015, p. 448–456.

\bibitem{10.5555/3045390.3045445}
W.~Liu, Y.~Wen, Z.~Yu, and M.~Yang, ``Large-margin softmax loss for
  convolutional neural networks,'' ser. ICML'16.\hskip 1em plus 0.5em minus
  0.4em\relax JMLR.org, 2016, p. 507–516.

\bibitem{8100196}
W.~Liu, Y.~Wen, Z.~Yu, M.~Li, B.~Raj, and L.~Song, ``Sphereface: Deep
  hypersphere embedding for face recognition,'' in \emph{2017 IEEE Conference
  on Computer Vision and Pattern Recognition (CVPR)}, 2017, pp. 6738--6746.

\bibitem{8331118}
F.~Wang, J.~Cheng, W.~Liu, and H.~Liu, ``Additive margin softmax for face
  verification,'' \emph{IEEE Signal Processing Letters}, vol.~25, no.~7, pp.
  926--930, 2018.

\bibitem{10.1007/978-3-319-46478-7_31}
Y.~Wen, K.~Zhang, Z.~Li, and Y.~Qiao, ``A discriminative feature learning
  approach for deep face recognition,'' in \emph{Computer Vision -- ECCV 2016},
  B.~Leibe, J.~Matas, N.~Sebe, and M.~Welling, Eds.\hskip 1em plus 0.5em minus
  0.4em\relax Cham: Springer International Publishing, 2016, pp. 499--515.

\bibitem{2019arXiv190301182C}
H.-Y. {Chen}, P.-H. {Wang}, C.-H. {Liu}, S.-C. {Chang}, J.-Y. {Pan}, Y.-T.
  {Chen}, W.~{Wei}, and D.-C. {Juan}, ``{Complement Objective Training},''
  \emph{arXiv e-prints}, p. arXiv:1903.01182, Mar. 2019.

\bibitem{7258343}
C.~Xu, C.~Lu, X.~Liang, J.~Gao, W.~Zheng, T.~Wang, and S.~Yan, ``Multi-loss
  regularized deep neural network,'' \emph{IEEE Transactions on Circuits and
  Systems for Video Technology}, vol.~26, no.~12, pp. 2273--2283, 2016.

\bibitem{9511448}
Y.~Li, Y.~Lu, B.~Chen, Z.~Zhang, J.~Li, G.~Lu, and D.~Zhang, ``Learning
  informative and discriminative features for facial expression recognition in
  the wild,'' \emph{IEEE Transactions on Circuits and Systems for Video
  Technology}, pp. 1--1, 2021.

\bibitem{2019arXiv190200220Z}
Q.~{Zhu} and R.~{Zhang}, ``{A Classification Supervised Auto-Encoder Based on
  Predefined Evenly-Distributed Class Centroids},'' \emph{arXiv e-prints}, p.
  arXiv:1902.00220, Feb. 2019.

\bibitem{8933403}
Q.~Zhu, P.~Zhang, Z.~Wang, and X.~Ye, ``A new loss function for cnn classifier
  based on predefined evenly-distributed class centroids,'' \emph{IEEE Access},
  vol.~8, pp. 10\,888--10\,895, 2020.

\bibitem{9444709}
H.~Hu, Y.~Yan, Q.~Zhu, and G.~Zheng, ``Generation and frame characteristics of
  predefined evenly-distributed class centroids for pattern classification,''
  \emph{IEEE Access}, vol.~9, pp. 113\,683--113\,691, 2021.

\bibitem{9184823}
S.~Yang, W.~Deng, M.~Wang, J.~Du, and J.~Hu, ``Orthogonality loss: Learning
  discriminative representations for face recognition,'' \emph{IEEE
  Transactions on Circuits and Systems for Video Technology}, vol.~31, no.~6,
  pp. 2301--2314, 2021.

\bibitem{8666165}
D.~Zhong and J.~Zhu, ``Centralized large margin cosine loss for open-set deep
  palmprint recognition,'' \emph{IEEE Transactions on Circuits and Systems for
  Video Technology}, vol.~30, no.~6, pp. 1559--1568, 2020.

\bibitem{8014803}
P.~Sermanet, C.~Lynch, J.~Hsu, and S.~Levine, ``Time-contrastive networks:
  Self-supervised learning from multi-view observation,'' in \emph{2017 IEEE
  Conference on Computer Vision and Pattern Recognition Workshops (CVPRW)},
  2017, pp. 486--487.

\bibitem{2021arXiv210303230Z}
J.~{Zbontar}, L.~{Jing}, I.~{Misra}, Y.~{LeCun}, and S.~{Deny}, ``{Barlow
  Twins: Self-Supervised Learning via Redundancy Reduction},'' \emph{arXiv
  e-prints}, p. arXiv:2103.03230, Mar. 2021.

\bibitem{2021arXiv210504906B}
A.~{Bardes}, J.~{Ponce}, and Y.~{LeCun}, ``{VICReg:
  Variance-Invariance-Covariance Regularization for Self-Supervised
  Learning},'' \emph{arXiv e-prints}, p. arXiv:2105.04906, May 2021.

\bibitem{2017arXiv170510284Y}
Y.~{Yuan}, K.~{Yang}, and C.~{Zhang}, ``{Feature Incay for Representation
  Regularization},'' \emph{arXiv e-prints}, p. arXiv:1705.10284, May 2017.

\bibitem{824819}
A.~Jain, R.~Duin, and J.~Mao, ``Statistical pattern recognition: a review,''
  \emph{IEEE Transactions on Pattern Analysis and Machine Intelligence},
  vol.~22, no.~1, pp. 4--37, 2000.

\bibitem{pytorch}
\BIBentryALTinterwordspacing
A.~Paszke, S.~Gross, S.~Chintala, G.~Chanan, E.~Yang, Z.~DeVito, Z.~Lin,
  A.~Desmaison, L.~Antiga, and A.~Lerer, ``Automatic differentiation in
  pytorch,'' in \emph{NIPS 2017 Workshop on Autodiff}, 2017. [Online].
  Available: \url{https://openreview.net/forum?id=BJJsrmfCZ}
\BIBentrySTDinterwordspacing

\bibitem{Krizhevsky2009LearningML}
A.~Krizhevsky, ``Learning multiple layers of features from tiny images,'' 2009.

\bibitem{5206848}
J.~Deng, W.~Dong, R.~Socher, L.-J. Li, K.~Li, and L.~Fei-Fei, ``Imagenet: A
  large-scale hierarchical image database,'' in \emph{2009 IEEE Conference on
  Computer Vision and Pattern Recognition}, 2009, pp. 248--255.

\bibitem{7025068}
H.-W. Ng and S.~Winkler, ``A data-driven approach to cleaning large face
  datasets,'' in \emph{2014 IEEE International Conference on Image Processing
  (ICIP)}, 2014, pp. 343--347.

\bibitem{8953658}
J.~Deng, J.~Guo, N.~Xue, and S.~Zafeiriou, ``Arcface: Additive angular margin
  loss for deep face recognition,'' in \emph{2019 IEEE/CVF Conference on
  Computer Vision and Pattern Recognition (CVPR)}, 2019, pp. 4685--4694.

\bibitem{app11146545}
\BIBentryALTinterwordspacing
T.~Kim, E.~Hong, and Y.~Choe, ``Deep morphological anomaly detection based on
  angular margin loss,'' \emph{Applied Sciences}, vol.~11, no.~14, 2021.
  [Online]. Available: \url{https://www.mdpi.com/2076-3417/11/14/6545}
\BIBentrySTDinterwordspacing

\bibitem{6296535}
L.~Deng, ``The mnist database of handwritten digit images for machine learning
  research [best of the web],'' \emph{IEEE Signal Processing Magazine},
  vol.~29, no.~6, pp. 141--142, 2012.

\bibitem{2016Densely}
G.~Huang, Z.~Liu, V.~Laurens, and K.~Q. Weinberger, ``Densely connected
  convolutional networks,'' in \emph{IEEE Computer Society}, 2016.

\bibitem{8578572}
M.~Sandler, A.~Howard, M.~Zhu, A.~Zhmoginov, and L.-C. Chen, ``Mobilenetv2:
  Inverted residuals and linear bottlenecks,'' in \emph{2018 IEEE/CVF
  Conference on Computer Vision and Pattern Recognition}, 2018, pp. 4510--4520.

\bibitem{2021An}
A.~Dosovitskiy, L.~Beyer, A.~Kolesnikov, D.~Weissenborn, X.~Zhai,
  T.~Unterthiner, M.~Dehghani, M.~Minderer, G.~Heigold, and S.~a. Gelly, ``An
  image is worth 16x16 words: Transformers for image recognition at scale,'' in
  \emph{International Conference on Learning Representations}, 2021.

\bibitem{2021arXiv210314030L}
Z.~{Liu}, Y.~{Lin}, Y.~{Cao}, H.~{Hu}, Y.~{Wei}, Z.~{Zhang}, S.~{Lin}, and
  B.~{Guo}, ``{Swin Transformer: Hierarchical Vision Transformer using Shifted
  Windows},'' \emph{arXiv e-prints}, p. arXiv:2103.14030, Mar. 2021.

\bibitem{2014Adam}
D.~Kingma and J.~Ba, ``Adam: A method for stochastic optimization,''
  \emph{Computer Science}, 2014.

\end{thebibliography}
%

\begin{IEEEbiography}[{\includegraphics[width=1in,height=1.25in,clip,keepaspectratio]{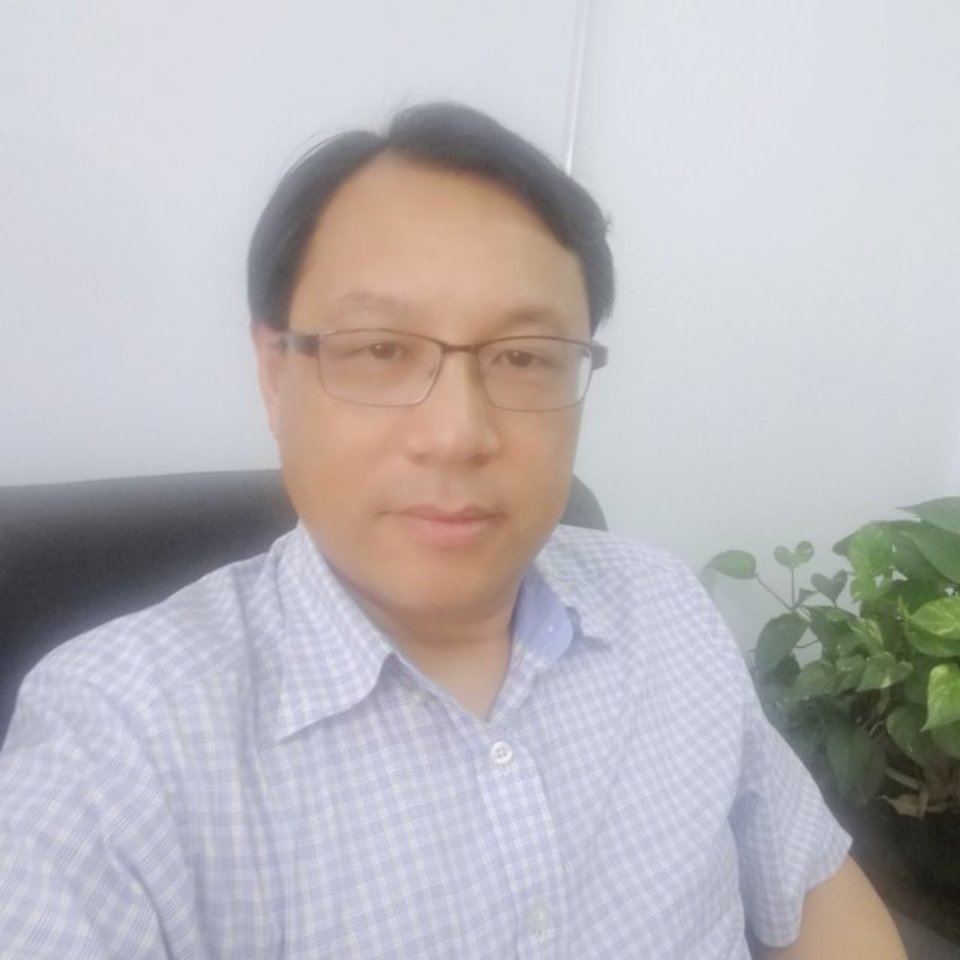}}]{QIUYU ZHU}
Received his Bachelor and Master degree from FUDAN University in 1985,and
Shanghai University of Science and Technology
in 1988 respectively. In 2006, he received his
PHD Degree in information and communication
engineering from Shanghai University. Now, he is
the professor in Shanghai University.His research
interests include image processing, computer vision, machine learning, smart city, computer application, etc. He is a coauthor of over 100
academic papers, and principal investigator for more than 10 governmental
funded research projects, more than 30 industrial research projects, many of
which have been widely applied. 
\end{IEEEbiography}

\begin{IEEEbiography}[{\includegraphics[width=1in,height=1.25in,clip,keepaspectratio]{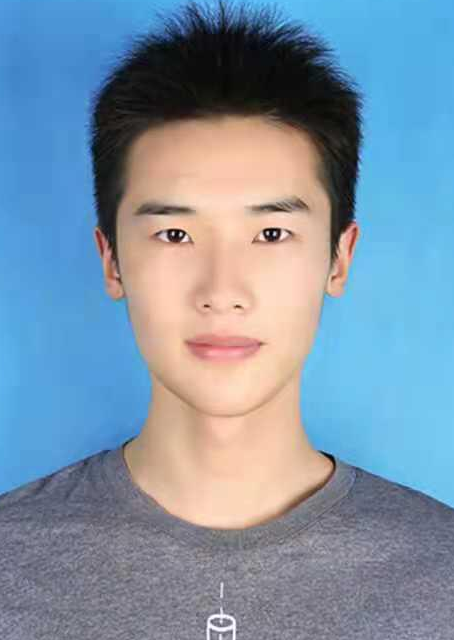}}]{XUEWEN ZU}
Received the bachelor's degree from the School of Communication
and Information Engineering, Shanghai University in 2020, he is currently pursuing the master's degree in the School of Communication and Information Engineering, Shanghai University. His research is computer vision.
\end{IEEEbiography}




\end{document}